\PassOptionsToPackage{backref=page}{hyperref}
\documentclass[11pt]{article}

\newif\ifarxiv
\arxivtrue 

\ifarxiv
    \usepackage{acl}
\else 
    \usepackage[review]{acl}
\fi

\usepackage{times}
\usepackage{latexsym}

\usepackage[T1]{fontenc}

\usepackage[utf8]{inputenc}

\usepackage{microtype}
\usepackage{sec/package}

\title{
Psychologically-Inspired Causal Prompts

}

\author{Zhiheng Lyu\thanks{\hspace{0.1cm} Equal contributions.}
\\
  The University of Hong Kong \\
  \texttt{zhlyu@cs.hku.hk} \\\And
  Zhijing Jin\samethanks \\
  MPI \& ETH Zürich \\
  \texttt{jinzhi@ethz.ch} \\\AND
  Mrinmaya Sachan \\
  ETH Zürich \\
  \texttt{msachan@ethz.ch} \\\And
  Rada Mihalcea \\
  University of Michigan \\
  \texttt{mihalcea@umich.edu} \\\And
  Bernhard Sch\"olkopf \\
  MPI for Intelligent Systems \\
  \texttt{bs@tue.mpg.de}\\
}

\begin{document}
\maketitle

\begin{abstract}
NLP datasets are richer than just input-output pairs; rather, they carry causal relations between the input and output variables. In this work, we take sentiment classification as an example and look into the causal relations between the review ($X$) and sentiment ($Y$). As psychology studies show that language can affect emotion, \textit{different psychological processes} are evoked when a person first makes a rating and then self-rationalizes their feeling in a review (where the sentiment causes the review, i.e., $Y\rightarrow X$), \textit{versus} first describes their experience, and weighs the pros and cons to give a final rating (where the review causes the sentiment, i.e., $X\rightarrow Y$). Furthermore, it is also a completely different psychological process if an annotator infers the original rating of the user by theory of mind (ToM) (where the review causes the rating, i.e., $X\xrightarrow{\mathrm{ToM}} Y$). In this paper, we verbalize these three causal mechanisms of human psychological processes of sentiment classification into three different causal prompts, and study (1) how differently they perform, and (2) what nature of sentiment classification data leads to agreement or diversity in the model responses elicited by the prompts.
We suggest future work raise awareness of different causal structures in NLP tasks.\footnote{Our code and data 
\ifarxiv
are at  \url{https://github.com/cogito233/psych-causal-prompt}.
\fi
}

\end{abstract}

\begin{figure}[t]
    \centering
    \includegraphics[width=\columnwidth]{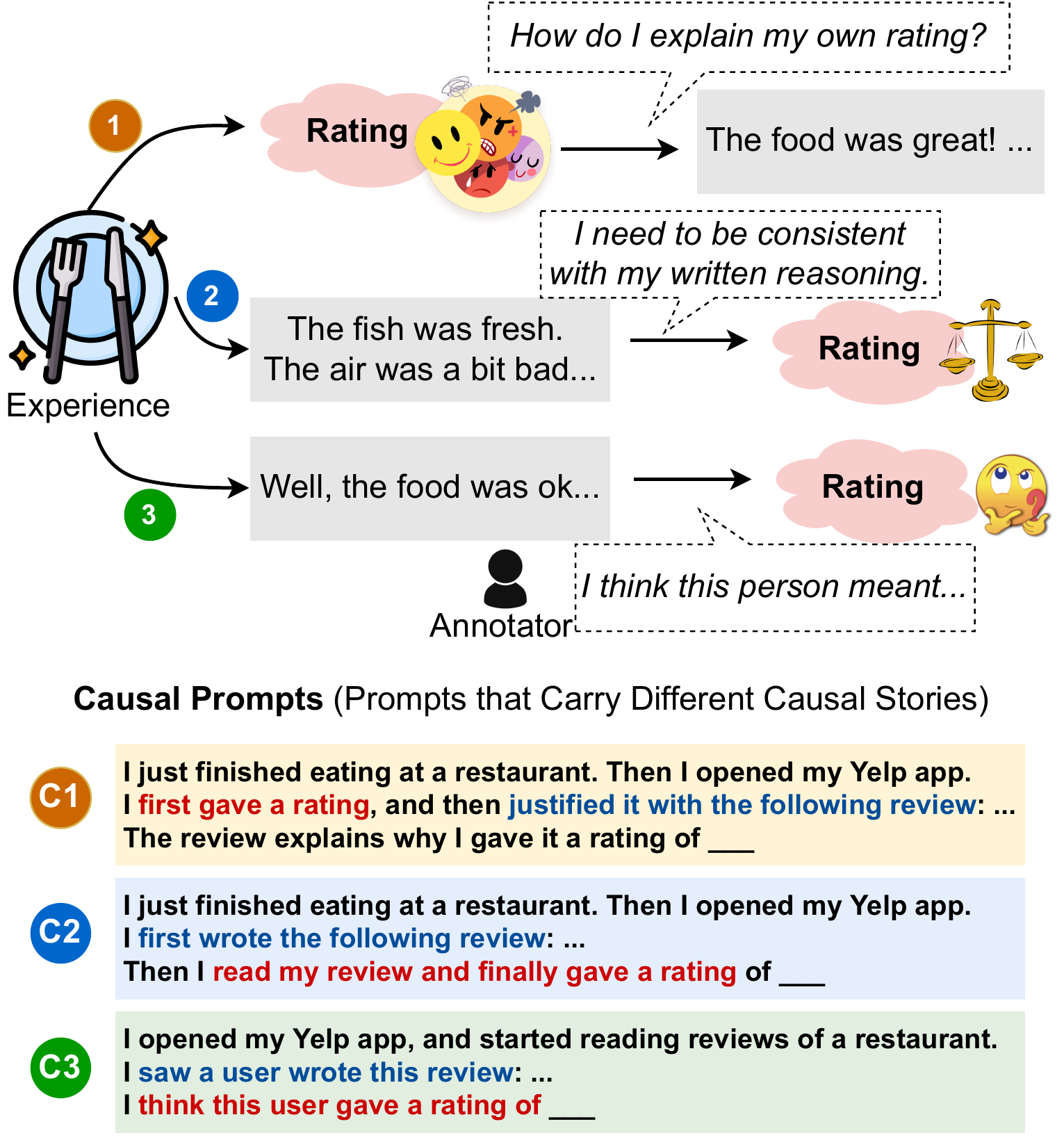}
    \caption{
    The review sentiment classification task might have different underlying psychological processes. We frame each of the three causal processes with a prompt, denoted as \textit{causal prompts} C1, C2, and C3.}
    \label{fig:intro}
    \vspace{-0.5cm}
\end{figure}

\section{Introduction}\label{sec:intro}
Most research on computational methods for NLP treats each task as input-output pairs. Take sentiment classification as an example. The common approach is to consider the review $X$ as input and the sentiment $Y$ as the output, from which the model tries to learn an optimal prediction $P(Y | X)$ \cite[see the surveys][]{liu2012survey,ravi2015survey,poria2020beneath}, making it a text classification task for which a standard text classifier can be finetuned \cite[\textit{inter alia}]{kim2014convolutional,yin-etal-2019-benchmarking}.

However, there can be rich underlying \textit{causal structures} between the input and output variables. For example, for the Yelp review data \citep{zhang2015character}, we list three different psychological processes that can lead to the review-sentiment pairs in \cref{fig:intro}: (C1) A user first makes a rating, and then self-rationalizes their feeling in a review (where an example psychological process is to justify their intuitive emotions), (C2) a user first describes their experience in a review, and then decides the rating (where the rating is often more rational, e.g., consistent with the facts in the review), and (C3) an annotator infers the original rating of a user by theory of mind (ToM) \cite{stich1983folk,frith2005theory}, which is possible for text classification tasks in general.

\begin{figure*}[t]
\minipage{0.33\textwidth}
  \includegraphics[width=\linewidth]{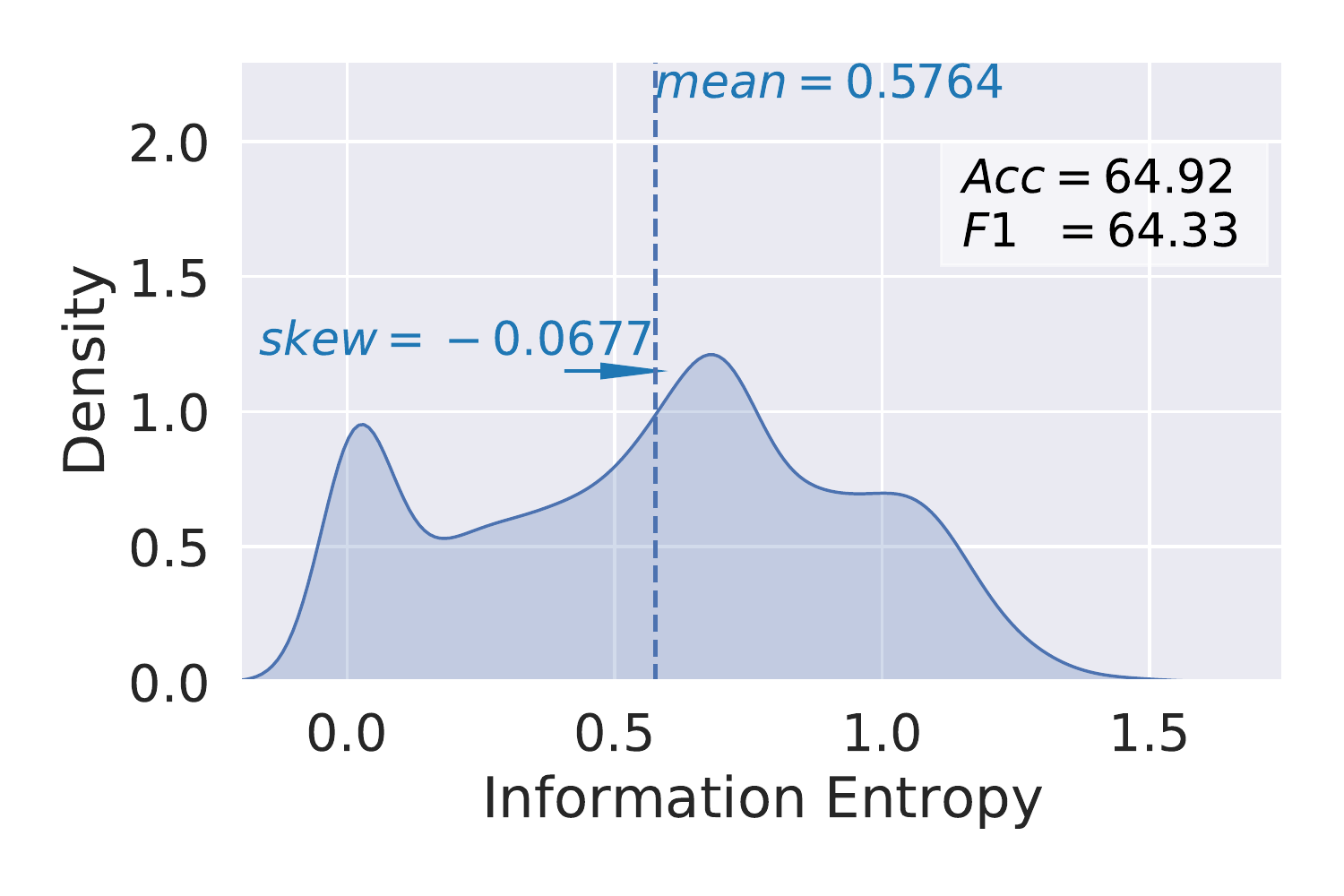}
\endminipage\hfill
\minipage{0.33\textwidth}
  \includegraphics[width=\linewidth]{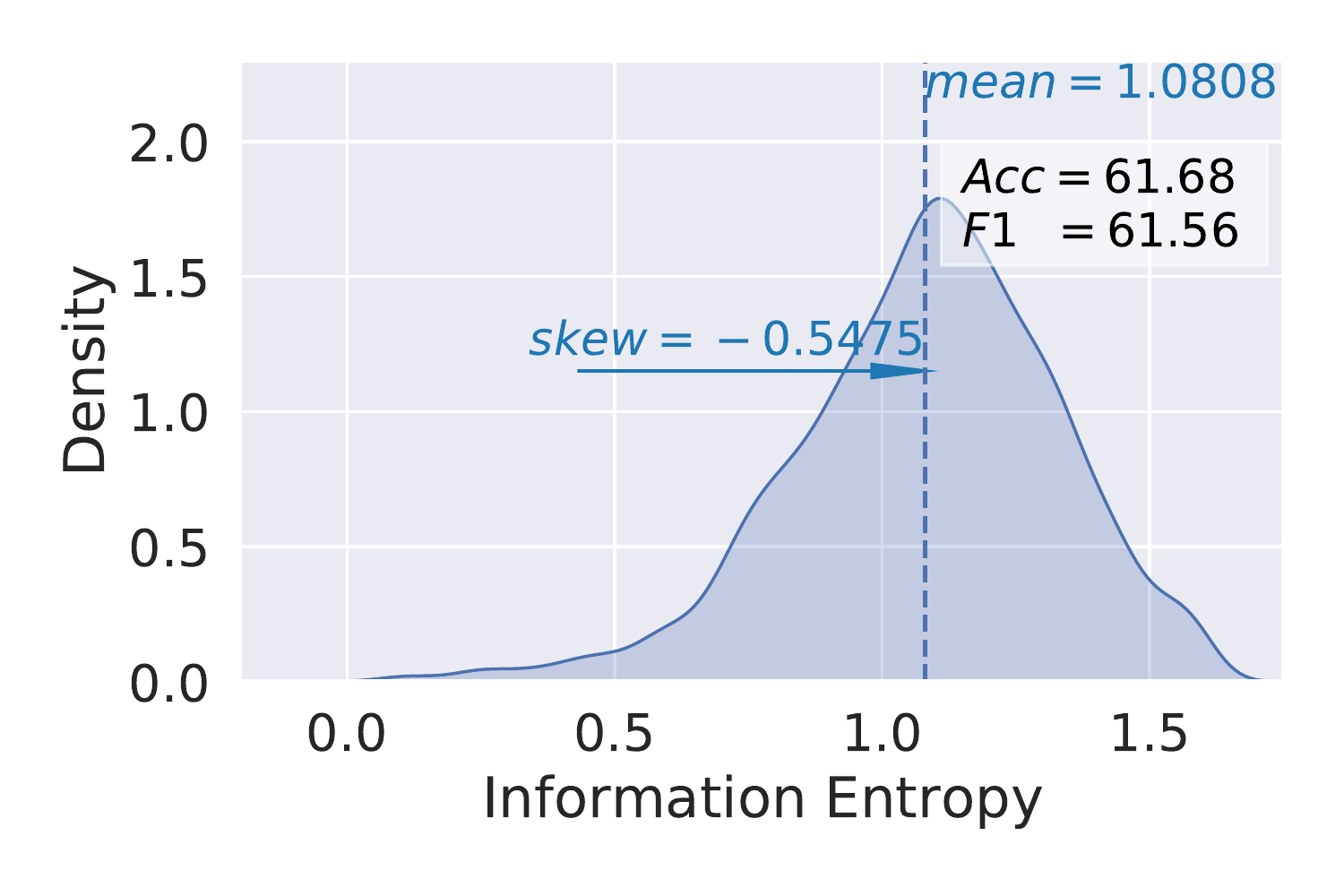}
\endminipage\hfill
\minipage{0.33\textwidth}
  \includegraphics[width=\linewidth]{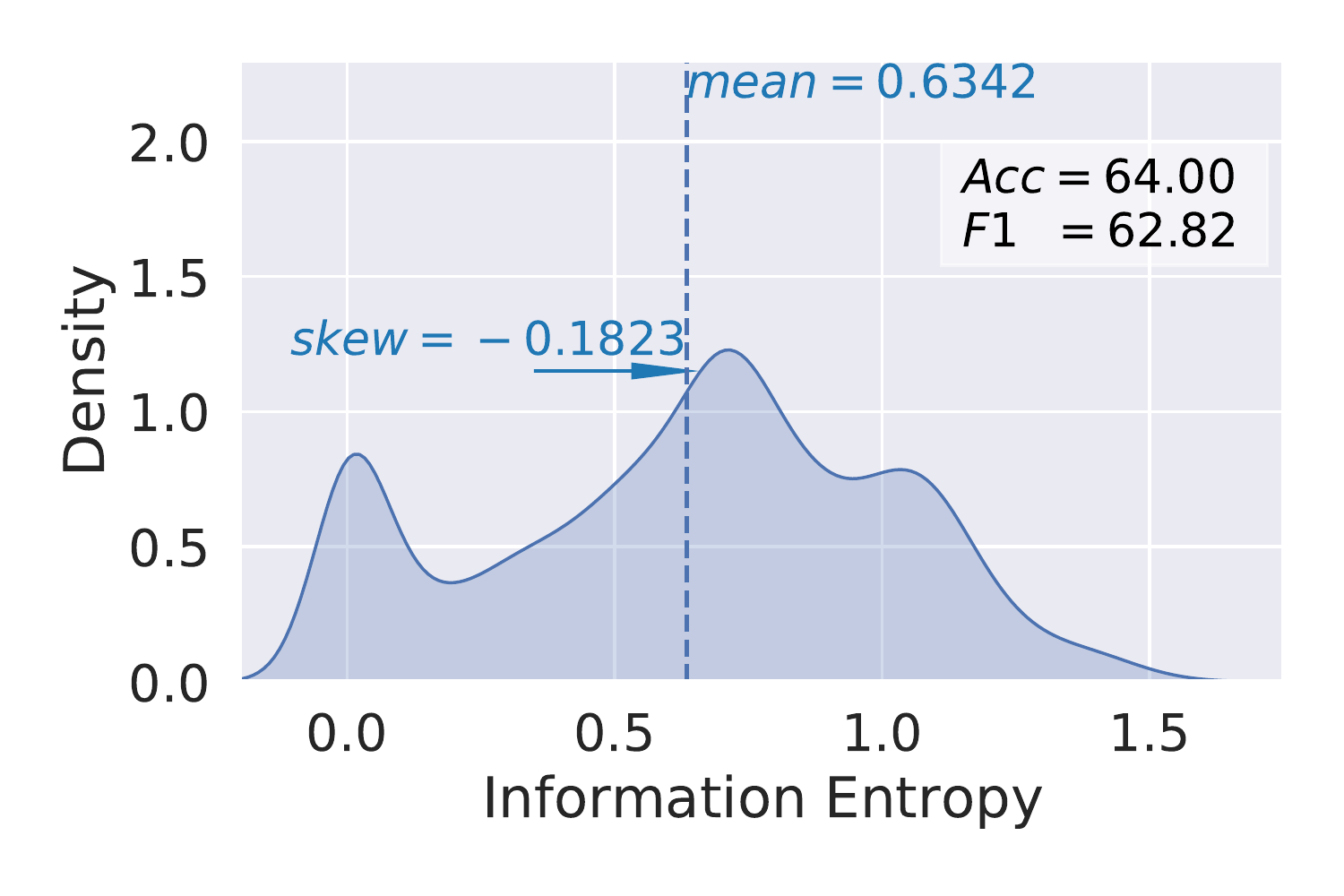}
\endminipage\hfill
\caption{
Performance of GPT3-Instruct \cite{ouyang2022instructGPT} induced by C1 (left), 2 (middle) and 3 (right) in terms of accuracy, weighted F1, and the distribution of information entropy for predictions across 5K English samples in the Amazon review test set \cite{keung-etal-2020-multilingual}.
  }
\label{fig:setup_distribution_amazon}
\end{figure*}

\begin{figure*}[t]
\minipage{0.33\textwidth}
  \includegraphics[width=\linewidth]{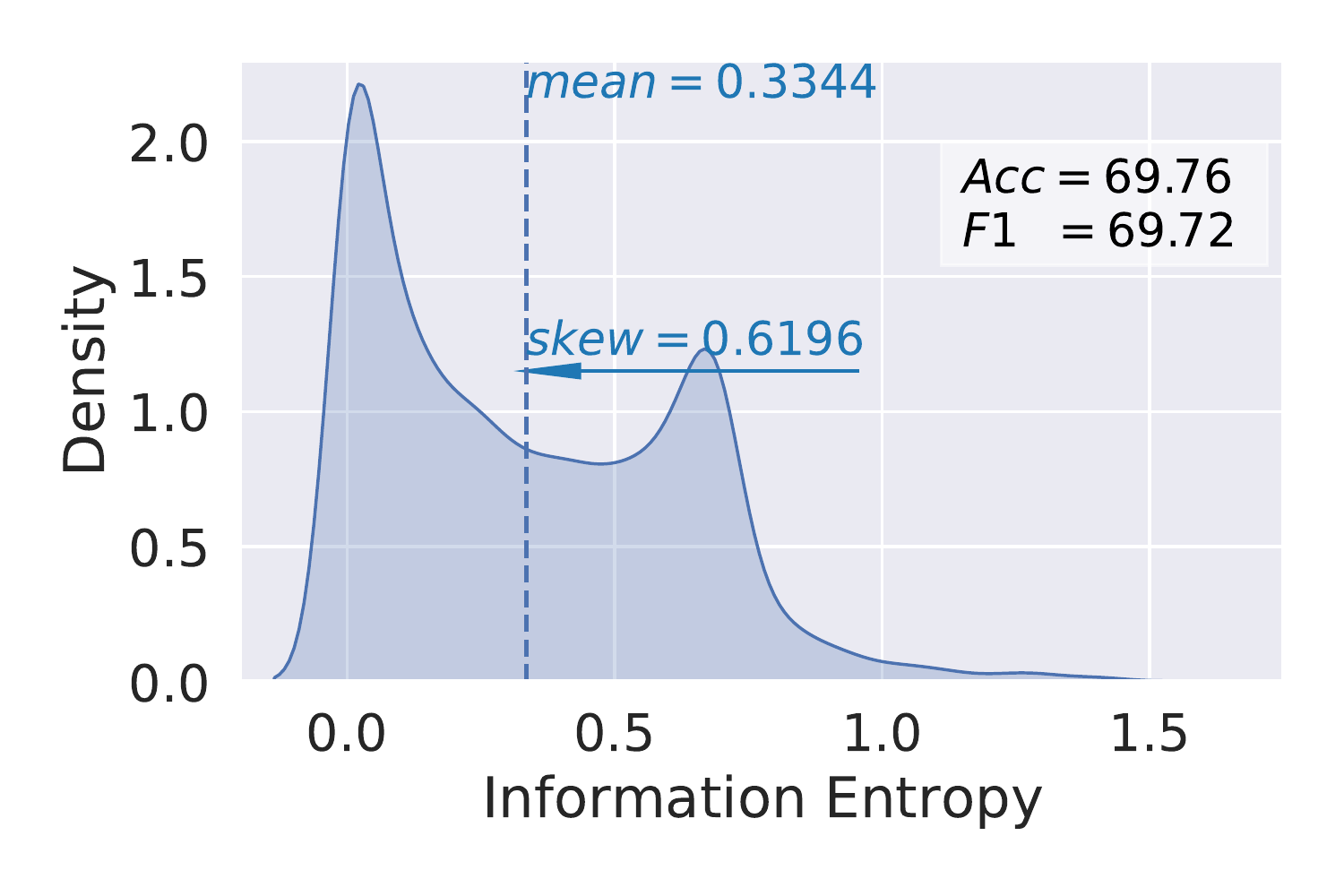}
\endminipage\hfill
\minipage{0.33\textwidth}
  \includegraphics[width=\linewidth]{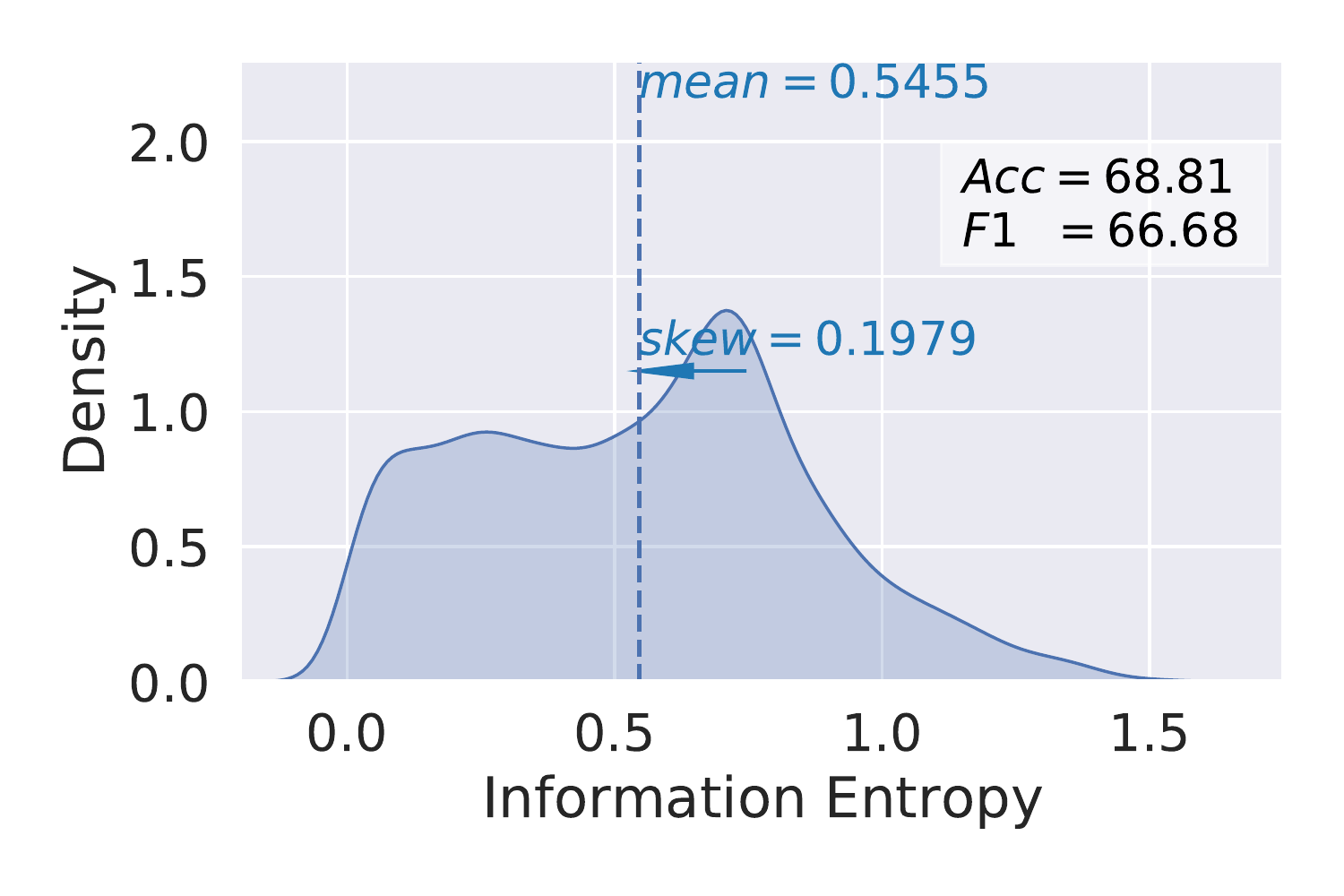}
\endminipage\hfill
\minipage{0.33\textwidth}
  \includegraphics[width=\linewidth]{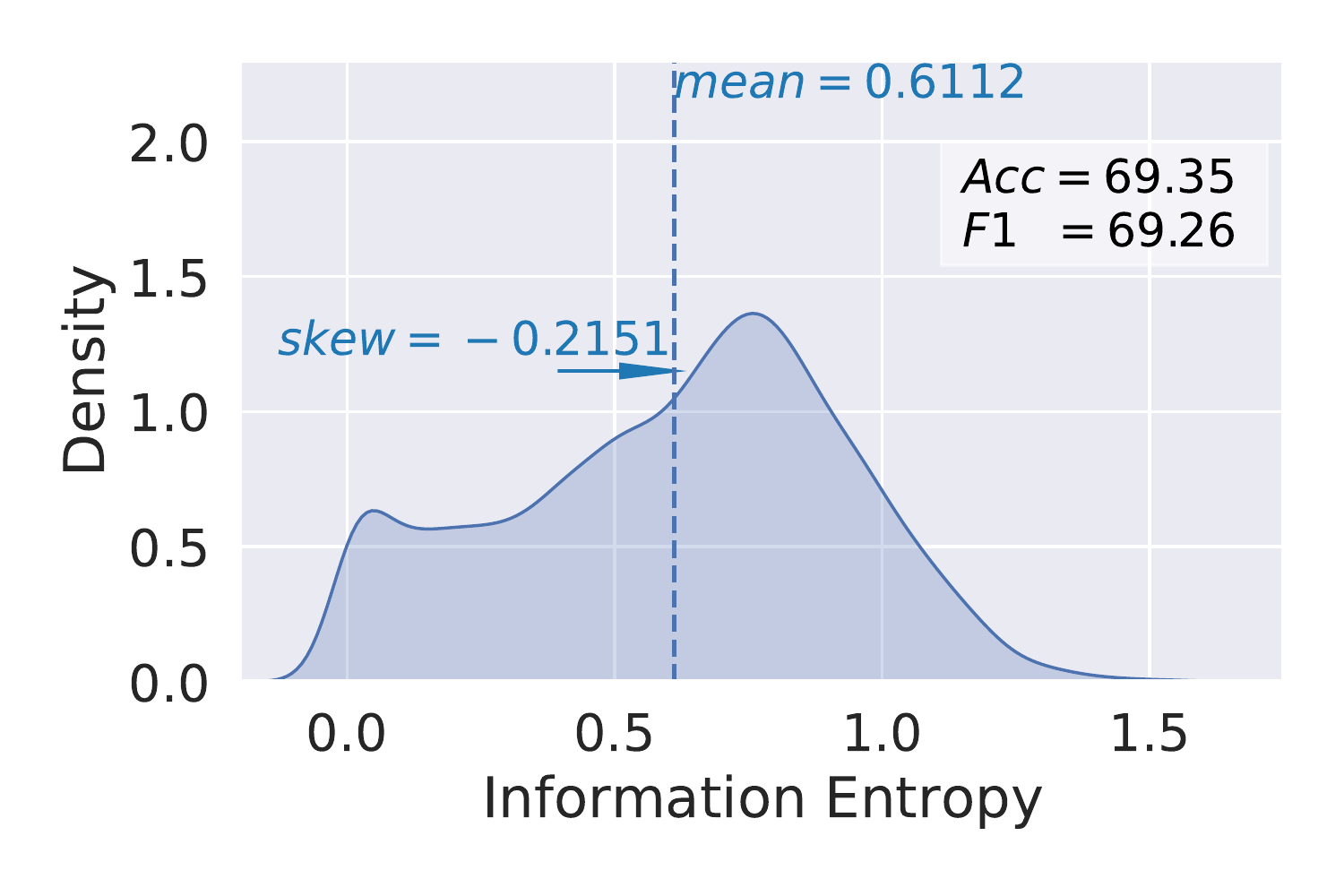}
\endminipage\hfill
\caption{
Performance of GPT3-Instruct \cite{ouyang2022instructGPT} induced by C1 (left), 2 (middle) and 3 (right) in terms of accuracy, weighted F1, and the distribution of information entropy for predictions across 10K samples in the Yelp-5 test set \citep{zhang2015character}.
  }
\label{fig:setup_distribution}
\end{figure*}

\ifarxiv
\begin{figure*}[t]
\minipage{0.33\textwidth}
  \includegraphics[width=\linewidth]{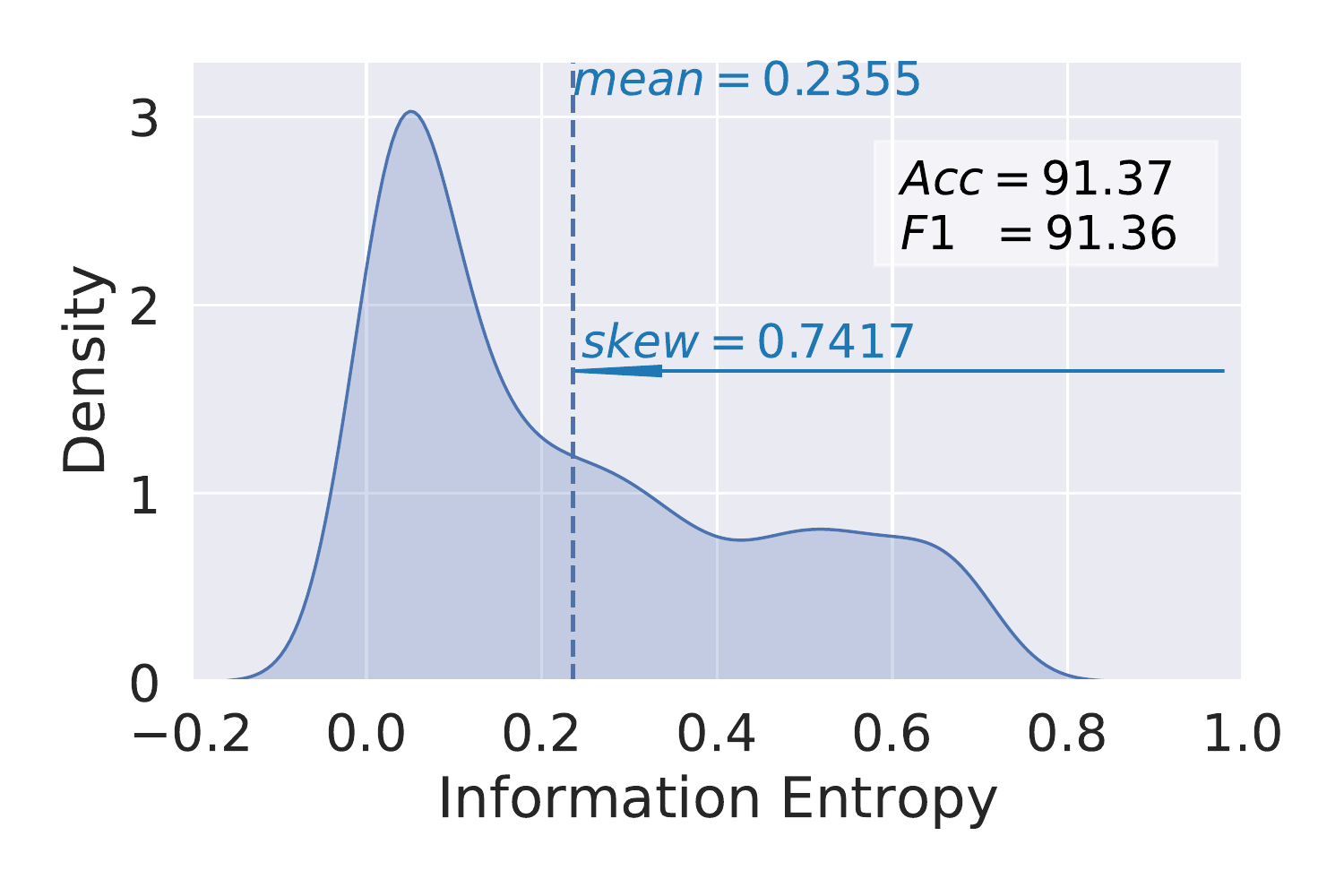}
\endminipage\hfill
\minipage{0.33\textwidth}
  \includegraphics[width=\linewidth]{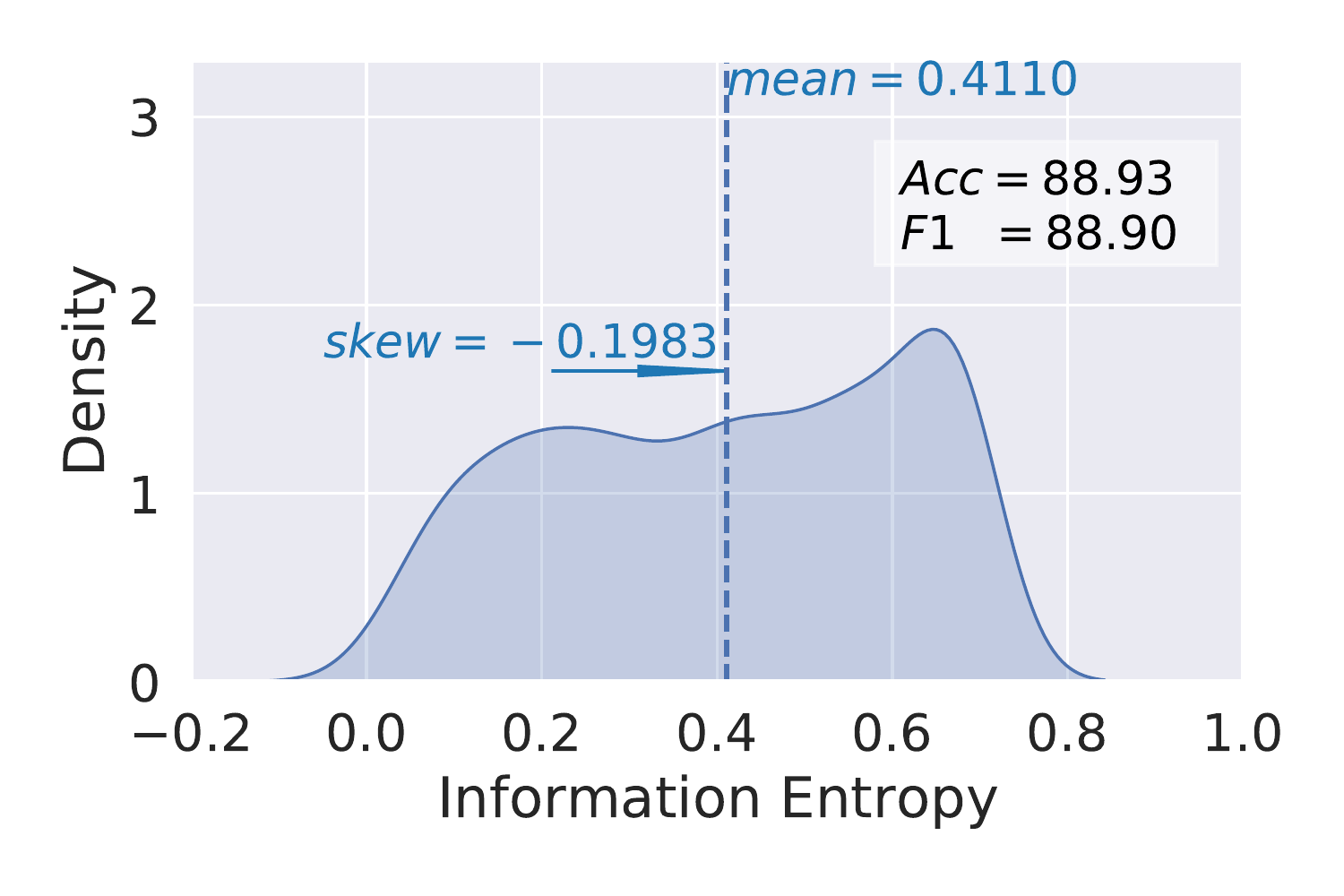}
\endminipage\hfill
\minipage{0.33\textwidth}
  \includegraphics[width=\linewidth]{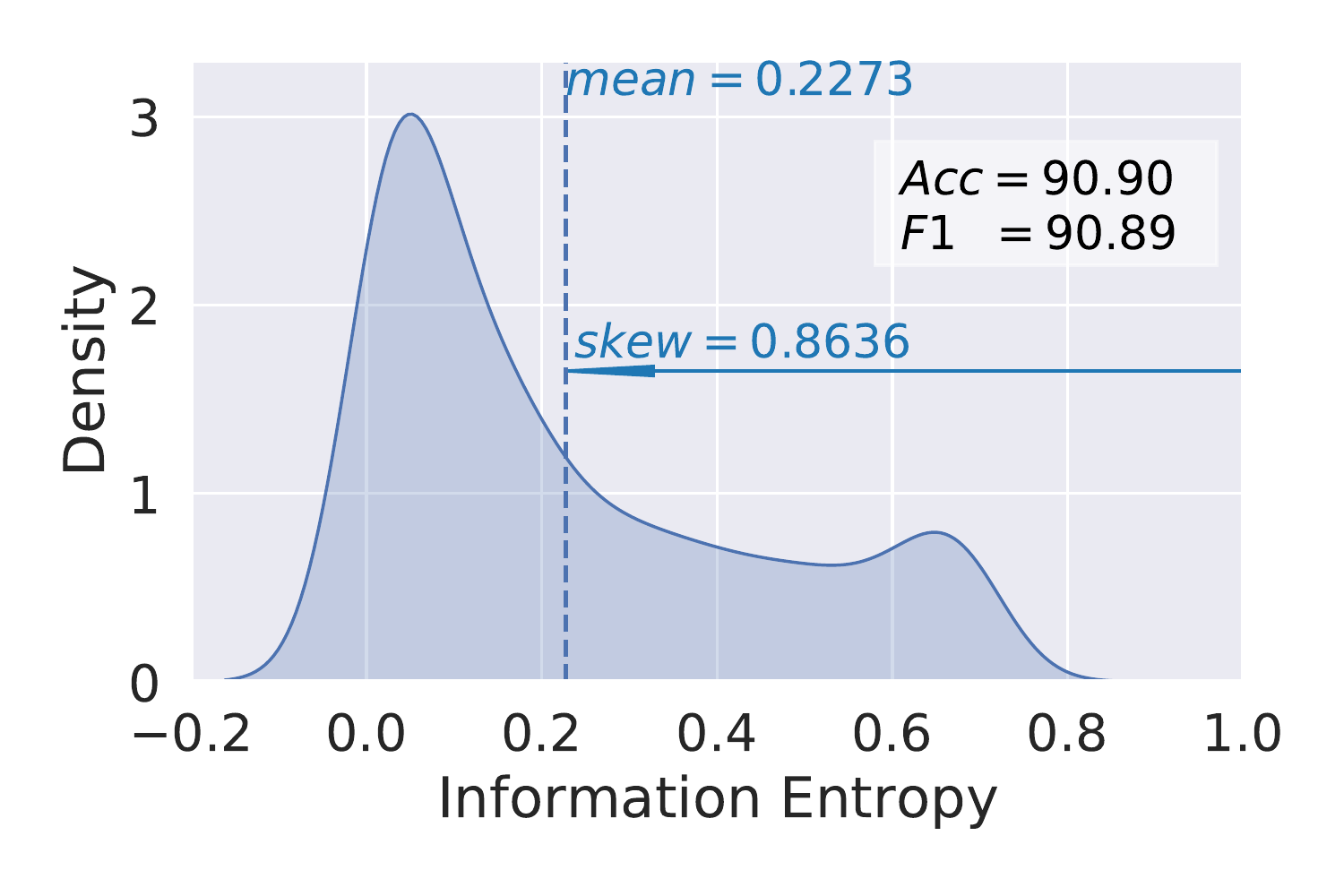}
\endminipage\hfill
\caption{
Performance of GPT3-Instruct \cite{ouyang2022instructGPT} induced by the C1 (left), 2 (middle) and 3 (right) in terms of accuracy, weighted F1, and the distribution of information entropy for predictions across the rotten-tomatoes dataset.
  }
\label{fig:setup_distribution_amazon}
\end{figure*}
\fi

In this short paper, we aim to provide insights to the question: \textit{how do such different psychological causal processes matter to NLP tasks?} We use sentiment classification as an example across this paper, for which we use psychology theories from affective science, create different \textit{causal prompts} that correspond to different psychological processes, and investigate two questions: (Q1) What different model behaviors do they elicit? And (Q2) What do model predictions elicited by the different causal prompts tell us about the nature of sentiment classification? 

We show that the three causal prompts lead to comparable model performance, but different certainty in their predictions. We further identify that in sentiment classification data, longer, more emotionally explicit reviews lead to more agreement across the causal prompts, whereas shorter, more implicit reviews lead to diverse predictions by the different causal prompts, which inspire us to reflect on the framing of the sentiment classification task, especially for reviews with more nuanced sentiment that could change by the underlying psychological process.

\section{Interdisciplinary Inspiration}
\myparagraph{Insights from Affective Science}
In the study of emotion, or affect science \cite{salovey2004emotional,barrett2006solving,feinstein2013lesion}, the interaction between language and affect has long been an important research area. For the (1) causal process, it is rather intuitive how emotion influences language, as the emotion people perceive influences how they communicate in the
moment \cite{barrett2006solving}. For the (2) causal process, affect science also have interesting discoveries of how language influences emotion. For example, the act of self-reporting on an emotional state is shown to change the physical reaction to an emotional situation \cite{kassam2013effects}. Such change from language to emotions can also be observed in functional magnetic neuroimaging \cite{satpute2013functional}.
For the (3) causal process, how a person infers the affect of another person falls into the study of ToM \cite{stich1983folk,frith2005theory}, which can be caused by both the stimulus and this person's capacity to infer the other person's mental state, causing perhaps inter-annotator variances from due to various understandings. 


\myparagraph{Causal and Anticausal Learning in NLP}\label{sec:formulation}
Our work also connects to the distinction of causal and anticausal learning in NLP \cite{jin-etal-2021-causal}, which distinguishes NLP tasks according to whether it learns to predict the cause from the effect, or the effect from cause. Applying to our work, the nature of the sentiment classification task changes if the underlying causal mechanisms flip. For example, if the C1 is the ground truth on a certain dataset, then a sentiment classifier conducts anticausal learning, whereas if C2 or C3 is the ground truth, then the sentiment classification task becomes causal learning.





\section{How Do Different Prompts Behave?}
In this section, we first explore Q1, namely what different behaviors the three different causal prompts can elicit.

\ifarxiv
\paragraph{Choosing the Causal Prompts}
Based on the three different psychological processes, we want to design three causal prompts for our follow-up analysis. We consider the following desiderata when coming up with the prompts: (1) Each prompt should unambiguously represent the causal process it correspond to. 
(2) The prompts should look like natural language, but not broken erraneous sentences. (3) The cross-causality comparison of the three prompts should be meaningful, which means that we need to account for other types of noises through which a prompt will lead to different responses of the LLMs.
Following these principles, we perform a careful prompt selection process detailed in \cref{appd:control}, with awareness of a broad literature on prompt design, including the survey \cite{liu2021pre} and recent work \cite{khashabi2022reframing,kojima2022large}. 

The three causal prompts we adapt are shown in \cref{fig:intro} (denoted as C1, C2, and C3), where we first ensure criteria (1) and (2), and then enforce criterion (3) to control the three prompts in terms of the structure, word choices, and lengths. To ensure that our three prompts have been well controlled, we also conduct additional analysis in \cref{appd:control}, such as how verbosity (i.e., keeping the meaning but changing the length) negatively affect three prompts in a similar way, and how perturbation in the prompts also uniformly decreases the performance of the three prompts. Our prompts have relatively comparable length, word choice, and naturalness.

\fi
\ifarxiv
\paragraph{Comparing the Prediction Distributions}
\fi
\ifarxiv
With the three carefully selected causal prompts, we observe their behavior 
\fi
on two commonly used English sentiment classification dataset, Yelp-5 \cite{zhang2015character}\footnote{\scriptsize \url{https://huggingface.co/datasets/Yelp_review_full}} and Amazon review \cite{keung-etal-2020-multilingual}.\footnote{\scriptsize \url{https://huggingface.co/datasets/amazon_reviews_multi}} The dataset has five labels corresponding to the 1 -- 5 ratings. We run the best available autoregressive LLM, GPT3-Instruct \cite{ouyang2022instructGPT}, on this fine-grained five-class classification, where which we can obtain richer information by comparing the predicted distribution over all five labels induced by different causal prompts. We report experimental details in 
\ifarxiv
\cref{appd:exp_details}.
\fi
\begin{table}[t]
    \centering \small
    \begin{tabular}{p{7.2cm}}
    \toprule
    \textit{\textbf{Reviews with High Agreement in Predictions}} \\
    \textbf{Review:} \textit{Terrible service!! Rude manager! Lost a customer for life. And the manager didn't care.} \\
    \textbf{Rating:} C1: {\color{red} {1 star}}. C2:{\color{red} {1 star}}. C3: {\color{red} {1 star}}. GT: {\color{red} {1 star}}.
    \\ \hline
    \textbf{Review:} \textit{OUTSTANDING COMMUNICATION, SERVICE AND QUALITY REPAIR!\newline I couldn't be happier!\newline FREE loan car! No, or partial deductible!\newline Only 8 business days to make my 2006 Solara look like new! I HIGHLY recommend them!\newline Tell 'em I sent you!!!} \\
    \textbf{Rating:} C1: {\color{blue} {5 stars}}. C2: {\color{blue} {5 stars}}. C3: {\color{blue} {5 stars}}. GT: {\color{blue} {5 stars}}.
    \\ \midrule \midrule
    \textit{\textbf{Reviews with High Diversity in Predictions}} \\
    \textbf{Review:} \textit{Confession.... I didn't eat here.\newline Truth.... I did walk in on Bill Clinton eating in the private dining room and was promptly told by the Secret Service to leave.\newline For the record... He looked like he was enjoying his meal.} \\
    \textbf{Rating:} C1: {\color{red} {1 star}}. C2: {\color{red} {1 star}}. C3: {\color{blue} {5 stars}}. GT: {\color{blue} {5 stars}}.
    \\ \hline
    \textbf{Review:} \textit{The waiting is too long!! They should expand their patio, offer a complementary drink if you wait is more then an hour! Who waits an hour to go out to eat! This is the reason i dont come here. She wanted to come here!} \\
    \textbf{Rating:} C1: {\color{red} {1 star}}. C2: {\color{red!40} {2 stars}}. C3: {\color{red} {1 star}}. GT: {\color{red!40} {2 stars}}.
    \\ 
    \bottomrule
    \end{tabular}
    \caption{Example reviews with high agreement or diversity in predictions by the three causal prompts (C1, C2, and C3). We also show the ground-truth label (GT).}
    \label{tab:review_examples}
    \vspace{-0.5cm}
\end{table}

\begin{table*}[t]
    \centering
    \small
    \setlength\tabcolsep{1pt}
    \begin{tabular}{lccccccccccccccc}
    \toprule
    & \multirow{2}{*}{\# Samples} & \multirow{2}{*}{\# Words/Sample} & \multicolumn{3}{c}{\# Opinion Words} & \multicolumn{5}{c}{Labels (\%)}
    & \multirow{2}{*}{Diversity}
    \\
    \cmidrule(lr){4-6} \cmidrule(lr){7-11}
    & && Pos & Neg & P$+$N & 1 & 2 & 3 & 4 & 5 \\\midrule
    
    Overall & 10,000 & 177.06 & 6.33{\tiny $\pm$5.54}  & 3.12{\tiny $\pm$3.59} & \lightredbox{9.44} & 20 & 20 & 20 & 20 & 20 & \bluebox{0.2337}\\ 
    Random (n=500) & 500 & 177.14 & 6.28{\tiny $\pm$5.31}  & 3.08{\tiny $\pm$3.58} & \lightredbox{9.36}  & 19 & 22 & 16 & 22 & 19 & \bluebox{0.2214}\\ \hline
    Same Correct: C1=C2=C3=GT & 5,994 & 184.17 & 6.51{\tiny $\pm$5.59}  & 3.28{\tiny $\pm$3.68} & \lightredbox{9.79}  & 25 & 17 & 17 & 16 & 23 & \bluebox{0.1679}\\ 
    Same Incorrect: C1=C2=C3$\ne$GT & 2,052 & 163.31 & 6.04{\tiny $\pm$5.28}  & 2.84{\tiny $\pm$3.44} & \lightredbox{8.89}  & 11 & 23 & 24 & 25 & 15 & \bluebox{0.2113}\\ 
    Diverse: Not (C1=C2=C3) & 1,954 & 169.68 & 6.04{\tiny $\pm$5.63}  & 2.90{\tiny $\pm$3.43} & \lightredbox{8.94}  & 12 & 24 & 22 & 24 & 15 & \bluebox{0.4591}\\ \midrule
    \textbf{Same Correct} \\
    \quad 10\% Data with Lowest Diversity & 599 & 231.90 & 6.39{\tiny $\pm$5.78}  & 5.10{\tiny $\pm$4.52} & \lightredbox{11.49}  & 53 & 9 & 23 & 3 & 10 & \bluebox{0.0167}\\ 
    \quad 10\% Data with Highest Diversity & 599 & 170.47 & 5.68{\tiny $\pm$5.21}  & 3.01{\tiny $\pm$3.39} & \lightredbox{8.69}  & 23 & 29 & 14 & 23 & 9 & \bluebox{0.4002}\\ \midrule
    \textbf{Same Incorrect} \\
    \quad 10\% Data with Lowest Diversity & 205 & 159.96 & 6.23{\tiny $\pm$4.85}  & 2.74{\tiny $\pm$2.45} & \lightredbox{8.97}  & 4 & 40 & 19 & 32 & 3 & \bluebox{0.0503}\\ 
    \quad 10\% Data with Highest Diversity & 205 & 178.51 & 5.74{\tiny $\pm$5.93}  & 3.25{\tiny $\pm$3.74} & \lightredbox{8.99}  & 17 & 24 & 30 & 11 & 16 & \bluebox{0.4163} \\ \midrule
    \textbf{Diverse} \\
    \quad Where C1$\ne$GT &  972 & 160.63 & 5.69{\tiny $\pm$5.58}  & 2.80{\tiny $\pm$3.30} & \lightredbox{8.49}  & 16 & 21 & 23 & 22 & 15 & \bluebox{0.4639}\\ 
    \quad Where C2$\ne$GT & 1,067 & 165.22 & 5.85{\tiny $\pm$5.43}  & 2.76{\tiny $\pm$3.27} & \lightredbox{8.61}  & 11 & 24 & 22 & 26 & 15 & \bluebox{0.4635} \\ 
    \quad Where C3$\ne$GT & 1,013 & 167.27 & 5.91{\tiny $\pm$5.61}  & 2.81{\tiny $\pm$3.38} & \lightredbox{8.72}  & 10 & 25 & 23 & 22 & 18 & \bluebox{0.4518} \\
    \bottomrule
    \end{tabular}
    \caption{
Statistics of subsets of data which induces different performance properties of the three causal prompts. We include subsets of test samples where all predictions all the same and correct (Same Correct), the same but incorrect (Same Incorrect), and diverse. For the two subsets with the same argmax predictions, we analyze the subsets that leads to the 10\% least diverse and most diverse model predictions. For the diverse argmax predictions, we analyze subsets that make each of the causal prompts (C1, C2, and C3) fail.
}
    \label{tab:res_subdata}
    \vspace{-0.3cm}
\end{table*}

We show the performance difference across the three prompts for the Amazon dataset in \cref{fig:setup_distribution_amazon} and Yelp dataset in \cref{fig:setup_distribution}.
For each prompt, we first report the 5-class accuracy and weighted F1 (on the top right corner of each figure), where C1 is slightly better than the other two.

Then, we also compare the information entropy of the three auto-completion process. Our intuition is that the entropy might show more subtle difference across the causal prompts in the log probability. A hypothesis in a recent work \cite{jin-etal-2021-causal} suggests that for review sentiment datasets, the users usually first make a rating and then write about it. This corresponds to C1 in our study, for which we observe noticeably more certainty in its predictions, through its low information entropy of 0.3344 on average, which is almost half of the other two causal prompts. Apart from the mean, we also calculate the skewness, where we can see that C1 is also skewed more towards lower entropy, with a skewness score of 0.6196.



\section{What Do Causal Prompts Tell Us about Our Data?}

After analyzing the properties of the LLM's responses to different prompts, we think from another angle --  what do different causal prompts tell us about our data?

We show our motivations in \cref{tab:review_examples}. We can see that the reviews with high agreement across the three causal processes tend to have some explicit signals, such as
clear opinion words, whereas the reviews that lead to large diversity in the predictions sometimes seem ambiguous or tricky for humans too.
For such nuanced reviews, the subtle difference among the different psychological processes elicited by the causal prompts may be more pronounced.


As a potential way for us to understand the nature of different sentiment classification data, e.g., where diversity might come from, we leverage GPT3-Instruct model \cite{ouyang2022instructGPT} and explore Yelp review-sentiment pairs where the responses by three causal prompts agree or diverge.


\myparagraph{Quantifying the Diversity}
For a given review $\bm{x}_i \in \bm{D}$, each model predicts a probability distribution $P(y | \bm{x}_i)$ of labels $y$ given the review. We quantify the diversity across the model responses to each two different causal prompts by the normalized total variation \cite{de1892serie,gibbs2002choosing}:
\setlength{\abovedisplayskip}{0pt}
\begin{align}
    d(P_1, P_2) = \frac{1}{2}| P_1(y | \bm{x}_i) - P_2(y | \bm{x}_i) | 
    ~,
\end{align}
where $P_1$ and $P_2$ are the distributions of model prediction by two causal prompts.
The metric $d: \mathcal{P} \times \mathcal{P} \rightarrow [0, 1]$ maps to a score between 0 and 1, where 0 is no diversity (complete agreement), and 1 is complete diversity, where two distributions have no overlapped areas.


\myparagraph{Quantifying Explicit Opinions}
Then, we look into properties of review text that might induce diverse opinions. One possible feature we can calculate is the number of explicit opinion words. As a preliminary exploration, we use a simple heuristic to match words in the review with the commonly used lists of 
positive\footnote{\scriptsize \url{https://ptrckprry.com/course/ssd/data/positive-words.txt}} and negative\footnote{\scriptsize \url{https://ptrckprry.com/course/ssd/data/negative-words.txt}} opinion words provided by \citet{hu2004mining,liu2005opinion}. 

\myparagraph{Findings} With the 10K test samples of Yelp-5, we analyze different subsets of it according to the prediction diversity. Some interesting properties of the data include the length of the text, the total number of opinion words, and the label distribution. For example, text with higher agreement (i.e., low diversity, the very light blue ones) tends to be those with more opinion words, and label distributions concentrated at certain labels. In contrast, text with higher prediction diversity (the darker blue ones) tends to have fewer opinion words and a relatively even label distribution slightly concentrated on ratings of 2 -- 4, as the decisions over 2 -- 4 are more ambiguous than the two extreme ends 1 and 5.

\myparagraph{Takeaways and Future Work}
Findings of this analysis inspire us to reflect on the framing of the sentiment classification task (and potentially many other NLP tasks). The key question is, how well defined are our tasks? Many tasks, although seemingly simple at first sight, might be subject to many subtle conditions. Future work needs to take more into considerations studies in other disciplines which provide rich insights on various potential causal processes underlying the input and output variables.

\section{Conclusion}
In this work, we proposed the concept of \textit{causal prompts}, and conducted analysis to report their performance, 
and reflect on the nature of sentiment classification.
We identified that in sentiment classification data, there exist many nuanced and ambiguous text that lead to diverse predictions that might correspond to different psychological interpretation of affect. Our findings raise an open question to the NLP community about how well defined our tasks are.

\section*{Limitations}
The main limitation of this short paper is that we mainly aim at a proof of concept, to make a suggestion that more care needs to be taken of the different causal processes we induce in the model. We put hedging on the experimental results as there could be other LLMs, other prompts and other datasets to try. In the scope of this short paper, we do not do a grid search over all combinations, but we encourage future work to explore with more solid and comprehensive set of experiments. Another limitation is that LLMs are not perfect. The way we are using LLMs, is based on the recent empirical success \cite{gpt3,ouyang2022instructGPT,chowdhery2022palm}, and a possibility that the correlations they capture can evoke psychologically-different causal processes between the review and the sentiment, as such framing might also be seen in its training data. In addition, these subtle differences that we explore might also touch upon another well known phenomenon, the measurement effect, where the act of measuring (reported sentiment) changes the properties of the observed (the actual affect). We encourage future work to explore more in depth. In addition, this study mainly covers the English language, and perhaps there are different phenomena for different language and cultures.

\section*{Ethical Considerations}
For data concerns and user privacy, we use a public dataset, and also the examples we quote are only for illustration purposes in this research paper. We remove any mentions of private persons' names in the examples.

For potential stakeholders and misuse, this study is in general to raise a new angle of thought for the sentiment classification task. A potential negative impact might be about the general task of sentiment classification, where a more accurate model might be used to mine user information, some negative uses of which include targetted marketing, surveilance, and fraud.

For the concerns over carbon footprints: Our study mainly uses the GPT-3 API. We try to control our carbon cost by subsampling the originally large test set to several thousand test 
samples, which are also relatively representative to show the trends.

\ifarxiv

\section*{Acknowledgment}
We thank Prof Erik C. Nook at the Department of Psychology at Princeton University for the discussions and references in affect science that helped us frame the idea of the paper.
This material is based in part upon works supported by the German Federal Ministry of Education and Research (BMBF): Tübingen AI Center, FKZ: 01IS18039B; by the Machine Learning Cluster of Excellence, EXC number 2064/1 – Project number 390727645; 
by the Precision Health Initiative at the University of Michigan; 
by the John Templeton Foundation (grant \#61156); by a Responsible AI grant by the Haslerstiftung; and an ETH Grant
(ETH-19 21-1).
Zhijing Jin is supported by PhD fellowships from the Future of Life Institute and Open Philanthropy, as well as the travel support from ELISE (GA no 951847) for the ELLIS program. We also thank OpenAI Researcher Access Program for granting our team credits to their API.

\section*{Author Contributions}\label{sec:contributions}
For the idea formulation, Zhijing Jin drew inspirations from her previous work \citet{jin-etal-2021-causal} and proposed to extend it to large language models.
During the internship of Zhiheng Lyu at ETH Zürich mentored by Zhijing, he first explored whether LLMs show a distinct fingerprint for causal and anticausal prompts in a workshop paper \cite{lyu2022can}. Then Zhijing got inspired by chats with psychology researchers in Affective Science that the phenomena of sentiment-primed text and text-primed sentiments can be grounded in actual psychology studies. Hence, we started to explore this paper's idea together.

\textit{Zhiheng Lyu} conducted all the experiments and analyses. He further proposed many further explorations, such as the analysis of the cross entropy.

\textit{Zhijing Jin} closely mentored the project and wrote the paper.

Professors \textit{Rada Mihalcea}, \textit{Mrinmaya Sachan} and \textit{Bernhard Schölkopf} supervised the project and improved the writing.

\fi

\bibliography{sec/refs_acl,sec/refs_this_paper,sec/refs_causality,sec/refs_zhijing,sec/refs_ai_safety,sec/refs_cogsci}
\bibliographystyle{acl_natbib}

\appendix

\section{Design of Causal Prompt}\label{appd:control}

\subsection{Desiderata of a Good Set of Three Causal Prompts}
Based on the three different psychological processes, we want to design three causal prompts for our follow-up analysis. We consider the following desiderata when coming up with the prompts: (1) Each prompt should unambiguously represent the causal process it corresponds to. 
(2) The prompts should be expressed in coherent natural language. (3) The cross-causality comparison of the three prompts should be meaningful, which means that we need to account for other types of noises through which a prompt will lead to different responses of the LLMs.

\subsection{Overall Prompt Selection Process}
Following these three principles, we perform a careful prompt selection process, with awareness of a broad literature on prompt design, including the survey \cite{liu2021pre} and recent work \cite{khashabi2022reframing,kojima2022large}. 

There are various ways to design a prompt \cite{liu2021pre}. The two main ways are discrete prompt design \cite[e.g.,][]{sorensen2022information,gao2021making} and continuous prompt design \cite[e.g.,][]{wu2022adversarial,zhu2022continual}. For continuous prompt design, we cannot use it for this work, because we need to keep the interpretability of our prompts so that humans can recognize and check the causal process it represents (unless future work comes up with an accurate embedding-to-causal process classifier). 
For discrete prompt design, we empirically evaluate some machine-generated paraphrases of our causal prompts, and find that purely machine-generated text often cannot fit criteria (1) and (2) together very well. For pure manual design, although the prompts can perfectly fit (1) and (2), some systematic human bias might be introduced \cite{cao2022can,kavumba2022prompt,elazar2021measuring}. 

To take the best from both worlds, we choose to do LLM-assisted manual framing of prompts. We use the most powerful autoregressive LLM that is available to us, the recent GPT3-Instruct \cite{ouyang2022instructGPT}, and we first describe the causal scenarios to it (``here is the process how a person ...''), and collect 10 different variations of how it rephrases each causal process, using the default temperature of 0.7 in its response generation.

Then based on the 10 different variations for each prompt, we filter out the ones that cannot meet criteria (1) and (2). Then, we reframe the candidates with minimal content word change but trying to follow tips of how to frame effective instructional prompts \cite{khashabi2022reframing}. As in \cref{fig:intro}, we frame the prompts to be a reasonable instruction, while at the same time also enforcing criterion (3), to control the three prompts as much as possible, in terms of structure, word choices, and length.





\subsection{Checking that the Prompts Are Well Controlled}
Although our main interest is the differences among prompts of different causal nature, there could be various noises in the prompt performance coming from other confounders. For example, two significant factors that could also affect the performance are
the length of the prompt and paraphrases with slight perturbations in words. 
Hence, in the following, we conduct some additional analyses: (1) how verbosity (i.e., keeping the meaning but changing the length of the prompt) negatively affect three prompts in a similar way, and (2) how perturbation in the prompts also uniformly decreases the performance of the three prompts.

Note that in the scope of our paper, we mainly consider these two sources of noise, but future work can explore other factors.


\paragraph{Comparing the Lengths.}

\begin{figure}[t]
\centering
  \includegraphics[width=0.8\linewidth]{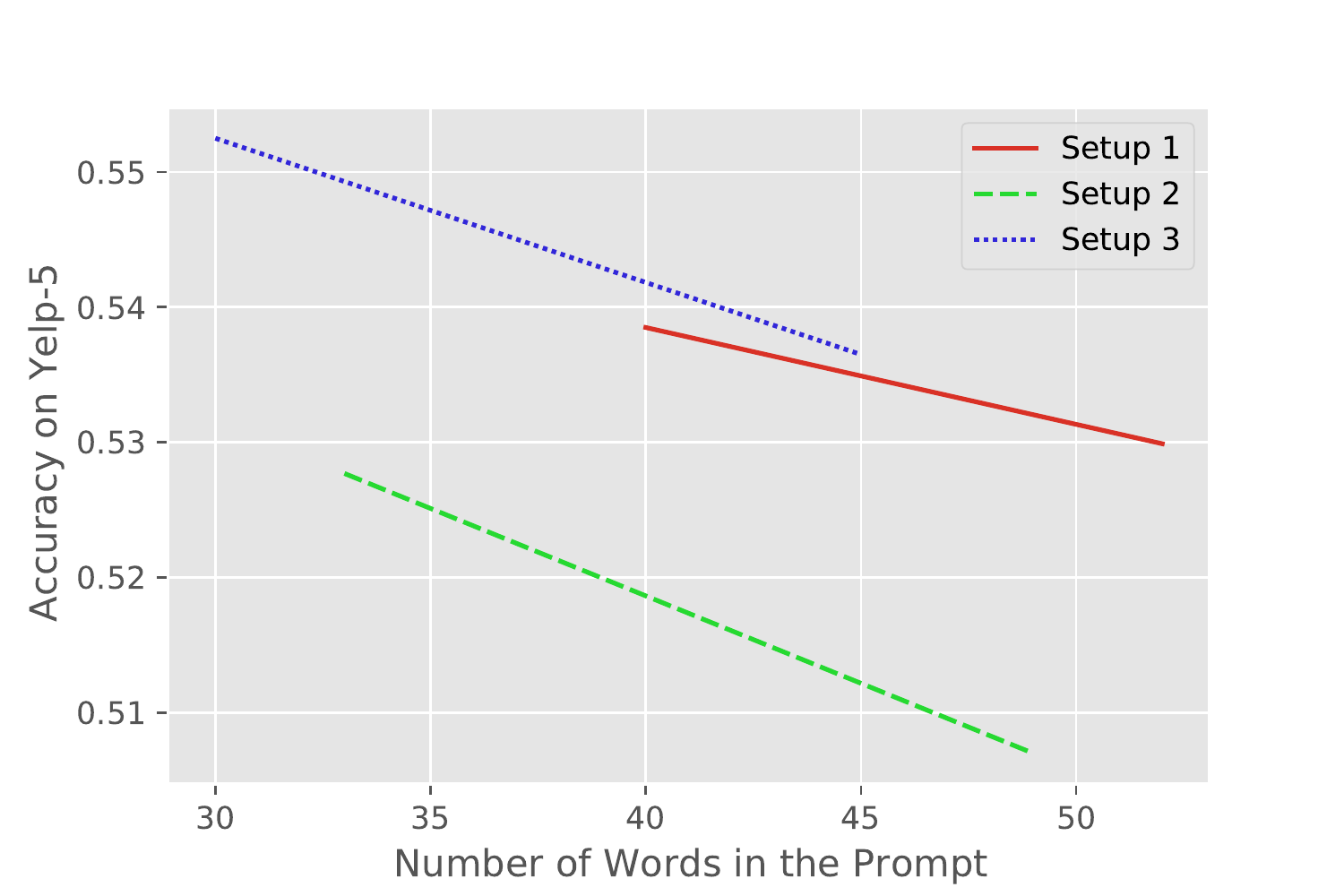}
  \caption{We confirm that the shorter version of the three causal prompts performs better than the long version, and the three short versions in \cref{fig:intro} are the best, controlled prompts we can come up with while fitting for the three desiderata. 
  }
    \label{tab:length_acc}
\end{figure}

For each causal prompt, we make pairs of long and short variations of it whose semantics and most word choices are the same, and only the verbosity is varied, as shown in \cref{tab:length_acc}.
We plot the performance of all three pairs of prompts in \cref{tab:length_acc}, where we can see that shorter prompts tend to have higher accuracy, and prompts of relatively the same lengths are comparable. Hence, we confirm that all the three short prompts we take, as shown in \cref{fig:intro}, make up a good set of causal prompts that relatively meet the last criterion.


\begin{table*}[t]
    \centering \small
    \begin{tabular}{p{15cm}llllll}
    \toprule
    \textbf{C1: Rating affects the review}
    \\
    \textit{\textbf{Short:}} I just finished eating at a restaurant. Then I opened my Yelp app. I first gave a rating, and then justified it by the following review: \texttt{[review text]} The review explains why I gave it a rating of \texttt{[Let GPTs complete]}
    \\
    \textit{\textbf{Long:}} I just finished eating at a restaurant. Then I opened my Yelp app. I first gave a rating in terms of 1 to 5 stars, and then explained why I gave the rating by the following review: \texttt{[review text]} The review is an explanation of why I rated it a \texttt{[Let GPTs complete]}
    \\ \midrule
    \textbf{C2: Review affects rating}
    \\
    \textit{\textbf{Short:}} I just finished eating at a restaurant. Then I opened my Yelp app. I first wrote the following review: \texttt{[review text]} Then I read my review and finally gave a rating of \texttt{[Let GPTs complete]}
    \\
    \textit{\textbf{Long:}} I just finished eating at a restaurant. Then I opened my Yelp app. I first wrote the following review: \texttt{[review text]} Then based on the review, I gave the rating in terms of 1 to 5 stars. I think this restaurant is worth a rating of \texttt{[Let GPTs complete]}
    \\ \midrule
    \textbf{C3: Another person guesses the rating from the review}
    \\
    \textit{\textbf{Short:}} I opened my Yelp app, and started reading reviews of a restaurant. I saw a user wrote this review: \texttt{[review text]} I think this user gave a rating of \texttt{[Let GPTs complete]} 
    \\
    \textit{\textbf{Long:}} I opened my Yelp app, and started to read some reviews of the restaurant that I wanted to try. I saw a user wrote this review: \texttt{[review text]} I think this user gave a rating (out of 1 to 5 stars) of \texttt{[Let GPTs complete]}
    \\ \bottomrule
    \end{tabular}
    \caption{Short and long prompts for each setup.}
    \label{tab:prompts_long_and_short}
\end{table*}

\begin{figure}
  \includegraphics[width=\linewidth]{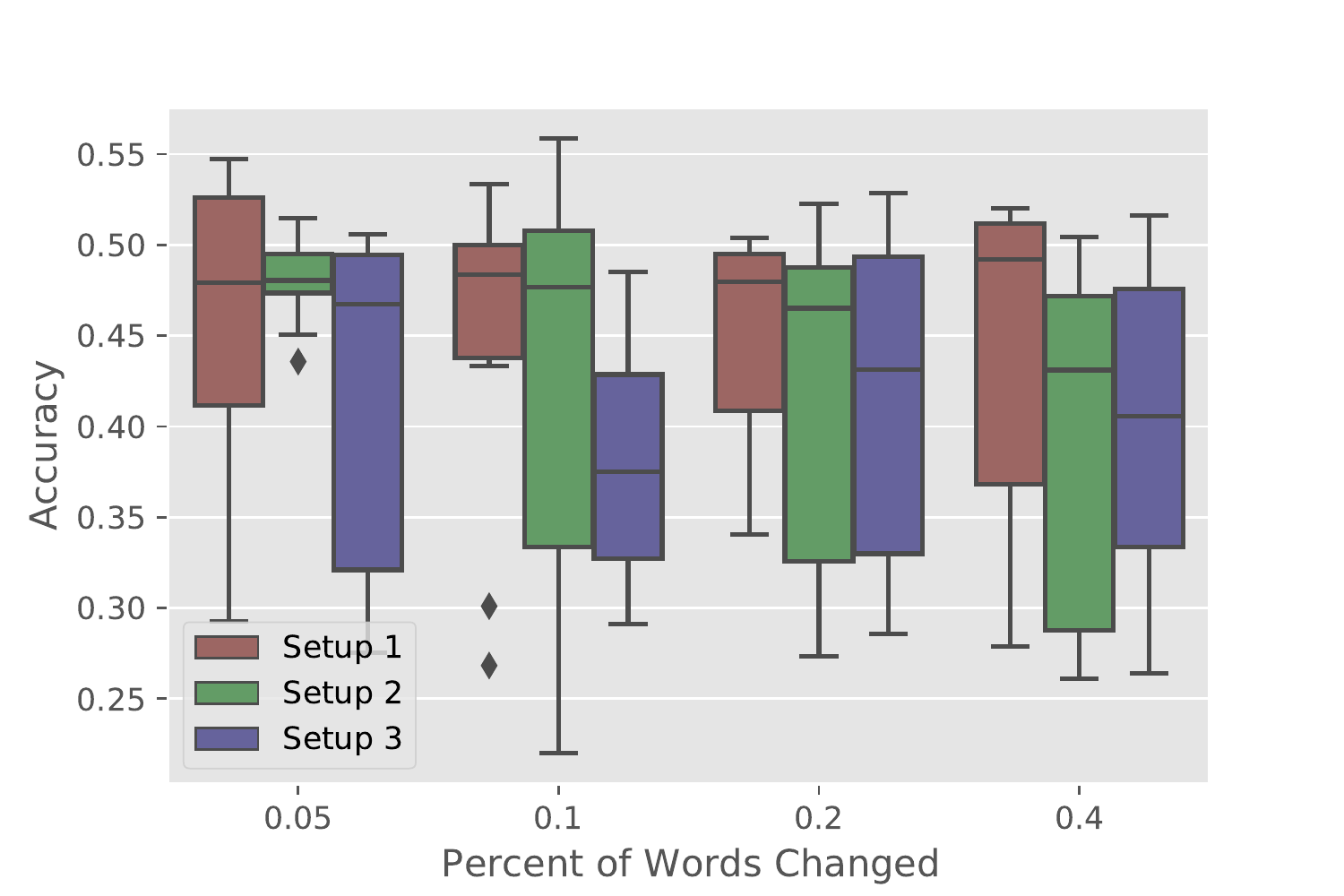}
  \includegraphics[width=\linewidth]{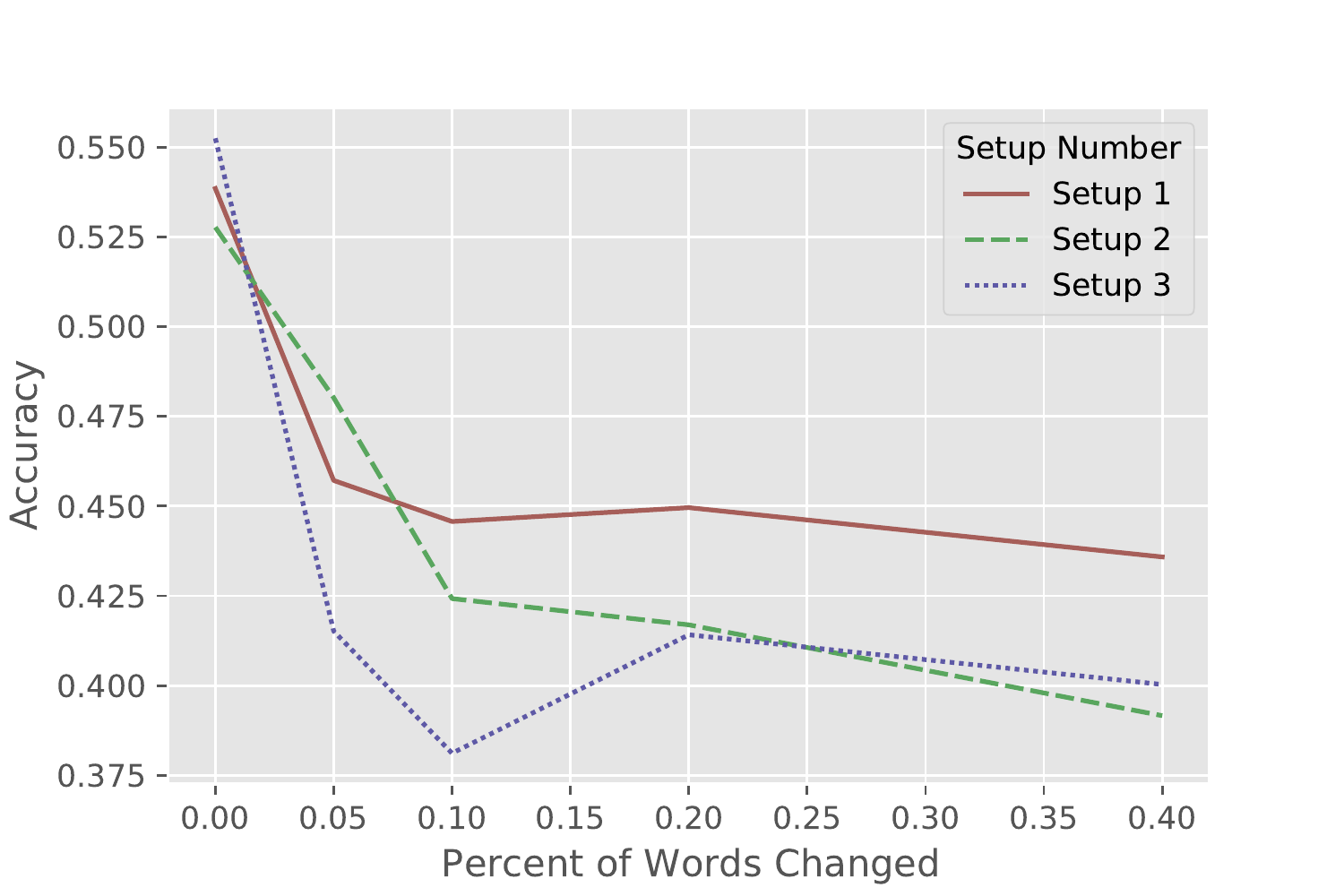}
  \caption{We confirm that our selection of prompts are relatively controlled in the sense that they are better than their perturbed versions, while meeting the three desiderata.
  }
    \label{tab:paraphrase_acc}
\end{figure}

\paragraph{Comparing with Slight Perturbations.}
We also check whether our prompts are controlled in terms of their relative performance to their variants with small perturbations. We generate a set of ten slight perturbations of the prompts using the easy data augmentation \cite{wei2019eda}, based on synonym
replacement, random insertion, random swap, and random deletion. We use the implementation in the TextAttack Python package \cite{morris-etal-2020-textattack}. 
We show their performance in \ref{tab:paraphrase_acc}, and confirm that the perturbations  decrease the performance.
Thus, we confirm
that, our prompts are the best performing one that can meet all three criteria, among all possible local perturbations with different perturbation strengths. 


\paragraph{Deciding Final Prompts.} 

The three causal prompts we adapt are shown in \cref{fig:intro} (denoted as C1, C2, and C3), where we first ensure criteria (1) and (2), and then enforce criterion (3) to control the three prompts in terms of the structure, word choices, and lengths. To ensure that our three prompts have been well controlled, we also conduct additional analysis in \cref{appd:control}, such as how verbosity (i.e., keeping the meaning but changing the length) negatively affect three prompts in a similar way, and how perturbation in the prompts also uniformly decreases the performance of the three prompts. Our prompts have relatively comparable length, word choice, and naturalness.

\section{Experimental Details}\label{appd:exp_details}
\subsection{Dataset}
We run our experiments on the widely used Yelp sentiment classification dataset \cite{zhang2015character}.
The original Yelp dataset has 650K samples in the training set, and 50K samples in the test set. The dataset has the same number of samples for each of the five labels 1 -- 5.

To reduce the computational and carbon costs, we downsample the dataset while keeping the label balance.
We select a random subset of 10K test samples from the original test set. We report all the performances across the paper using these 10K test samples.

\subsection{Model Details}

For the use of GPT3-Instruct \cite{ouyang2022instructGPT}, we use the API provided by OpenAI,\footnote{\scriptsize \url{https://openai.com/api/}} and its newest model at the point we publish this paper, ``\texttt{davinci-text-002}'' with 175B parameters.

To make our results reproducible, we set the temperature to be 0 for all uses of OpenAI APIs. At inference time, we obtain the top 5 candidates with their log probabilities. For the probability of each label 1, 2, \dots, 5, we add up the log probabilities of their English word form (e.g., ``one'') and the Arabic number (e.g., ``1'').

Since we do not finetune the model, but only call the API to run it in inference mode, the energy cost is relatively low, with around 20-30K single API calls, costing around 400 USD budget.

\subsection{Implementation Details}
Since models might have a different existing preference over different labels (e.g., more like to predict a neutral score such as ``3''), we use a small portion, 1K samples, of the training set to learn an additional parameter $\bm{\lambda} := (\lambda_1, \dots, \lambda_5)$, which adjusts the model's prediction $P(Y=y_i | X)$ by a scalar $\lambda_i$.
We learn $\bm{\lambda}$ by optimizing 
$L_1(P(\arg\max_y (\lambda_y P(Y=y | X=x))),P(Y))$, for a loss function $\mathcal{L}$. We list the learned values of $\bm{\lambda}$ in \cref{tab:prior_prob}.








\begin{table}[h]
    \centering
    \small
    \begin{tabular}{lcccccc}
    \toprule
    \multirow{2}{*}{Setup} & \multicolumn{5}{c}{Prior on Labels}
    \\ \cline{2-6}
    & $\lambda_1$ & $\lambda_2$
    & $\lambda_3$ & $\lambda_4$ & $\lambda_5$
    \\ \midrule
C1
    & 0.73 & 0.22 & 0.03 & 0.004 & 0.01\\
C2
    & 0.07 & 0.32 & 0.44 & 0.07 & 0.10\\
C3
    & 0.13 & 0.39 & 0.09 & 0.08 & 0.31 \\
    \bottomrule
    \end{tabular}
    \caption{
    The learned scaling factor for the prediction of the labels 1 -- 5 for GPT3-Instruct.
    }
    \label{tab:prior_prob}
\end{table}
\subsection{Information Entropy}

Given the predicted distribution $P(Y | X=x_i)$ for each sample $x_i\in \bm{D}$ in the test set $x_i$, we calculate The information entropy of this distribution by
\begin{align}
    H(X)=-\sum P(Y| x_i)\log{P(Y| x_i)}
    ~,
\end{align}
which is an indicator of the model's uncertainty over all possible labels.
Note that entropy is closely related to perplexity, which is calculated as $\mathrm{PPL}(P)=2^{H(P)}$.


\subsection{Explicit Opinion Word Count}
A simple indicator for sentiment is the set of words that carry strong opinions.
To count the occurrences of opinion words, we first tokenize the review by the GPT2-large tokenizer provided by the \texttt{transformers} library \cite{wolf2019transformers}, then we remove all symbols that are not regular English letters, Arabic numbers, and common symbols such as ``-,'' before matching with the given opinion word lists.


\section{More Examples of Reviews with Agreed or Diverse Predictions}
See \cref{tab:review_more_examples} for more examples of reviews with high agreement or diversity in predictions.

\begin{table}[t]
    \centering \small
    \begin{tabular}{p{7.2cm}}
    \toprule
    \textit{\textbf{Reviews with High Agreement in Predictions}} \\
    \textbf{Review:} \textit{Terrible service!! Rude manager! Lost a customer for life. And the manager didn't care.} \\
    \textbf{Rating:} C1: 1 star. C2: 1 star. C3: 1 star. GT: 1 star.\\ \\
    \textbf{Review:} \textit{The first time couple of times I went there they were closer to four stars, but this last trip put them at a three star. The combo Kung pao was just average. I really hate meat that is mushy, and that's what the beef was. This is just your average Chinese joint with a nice decor.} \\
    \textbf{Rating:} C1: 3 stars. C2: 3 stars. C3: 3 stars. GT: 3 stars.\\ \\
    \textbf{Review:} \textit{OUTSTANDING COMMUNICATION, SERVICE AND QUALITY REPAIR!\newline I couldn't be happier!\newline FREE loan car! No, or partial deductible!\newline Only 8 business days to make my 2006 Solara look like new! I HIGHLY recommend them!\newline Tell 'em I sent you!!!} \\
    \textbf{Rating:} C1: 5 stars. C2: 5 stars. C3: 5 stars. GT: 5 stars.\\ \\ 
    \textbf{Review:} \textit{The sushi is really good, but it's kinda pricy. The customer service is kinda weird too. Overall it's a decent place to go though.} \\
    \textbf{Rating:} C1: 3 stars. C2: 3 stars. C3: 3 stars. GT: 4 stars.\\ \\\midrule
    \textit{\textbf{Reviews with High Diversity in Predictions}} \\
    \textbf{Review:} \textit{Confession.... I didn't eat here.\newline Truth.... I did walk in on Bill Clinton eating in the private dining room and was promptly told by the Secret Service to leave.\newline For the record... He looked like he was enjoying his meal.} \\
    \textbf{Rating:} C1: 1 star. C2: 1 star. C3: 5 stars. GT: 5 stars.
    \\ \\
    \textbf{Review:} \textit{"Has the potential to be awesome, but unfortunately extremely disappointing.  Bartenders were awful, not nice, and had no idea what they were doing. Medium rare burger came out well done, with cold fries. It wasn't even busy in there and the whole staff looked flustered. Will not be returning.} \\
    \textbf{Rating:} C1: 2 stars. C2: 1 star. C3: 1 star. GT: 1 star.\\ \\
    \textbf{Review:} \textit{The waiting is too long!! They should expand their patio, offer a complementary drink if you wait is more then an hour! Who waits an hour to go out to eat! This is the reason i dont come here. She wanted to come here!} \\
    \textbf{Rating:} C1: 1 star. C2: 2 stars. C3: 1 star. GT: 2 stars.
    \\ \\
    \bottomrule
    \end{tabular}
    \caption{Example reviews with high agreement or diversity in predictions by the three causal prompts (C1, C2, and C3). We also show the ground-truth label (GT).}
    \label{tab:review_more_examples}
\end{table}

\section{Extended Analysis}
\subsection{Complete Plot of Opinion Words and Diversity}
\begin{figure}[t]

\centering
  \includegraphics[width=0.75\linewidth]{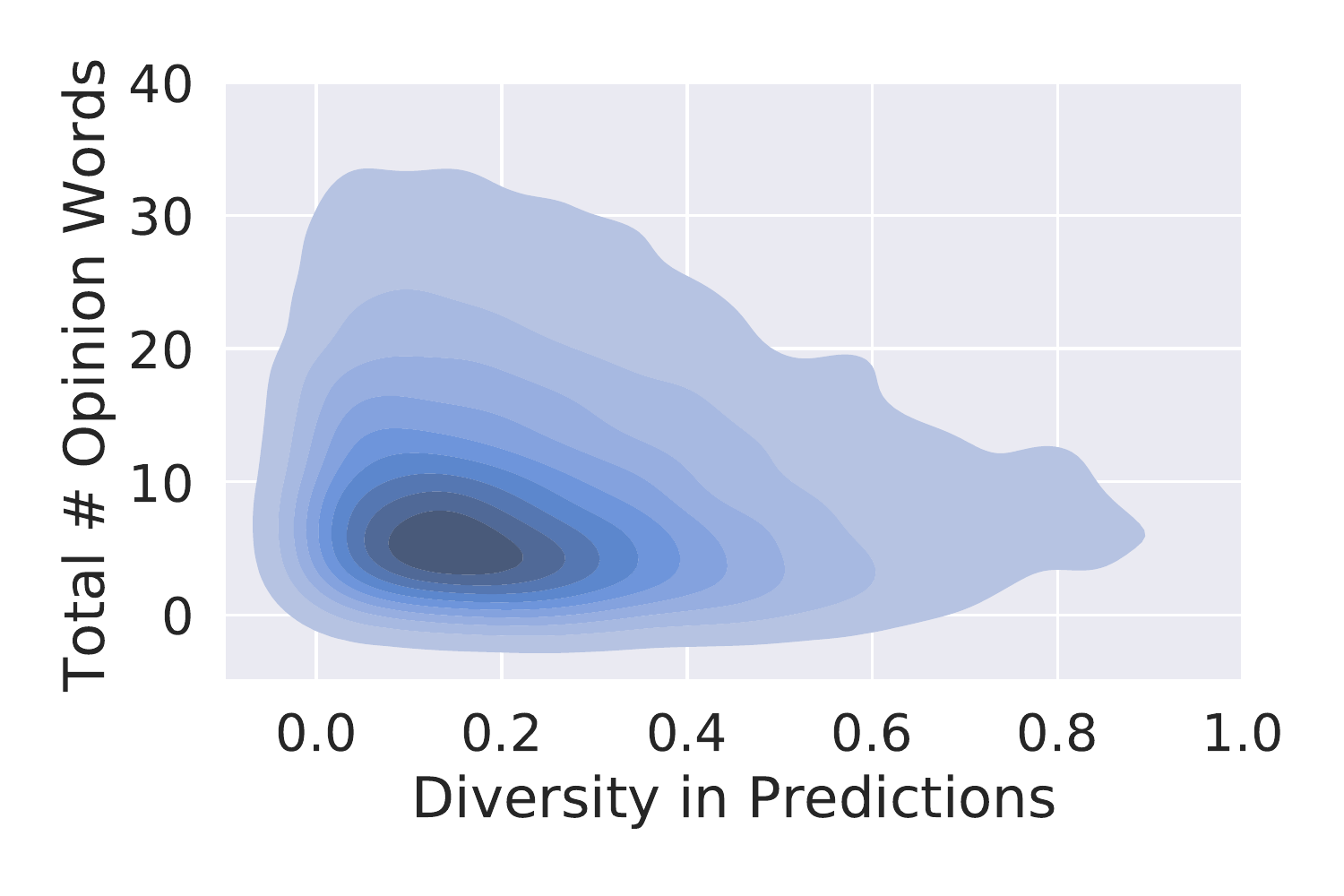}
\caption{
  The joint distribution of the total number of opinion words and the diversity in predictions. 
  }
\label{fig:pos_neg_distribution}

\end{figure}

For all the 10K test samples of Yelp, we plot all pairs of the total number of opinion words and the prediction diversity among all three causal prompts) in \cref{fig:pos_neg_distribution}.

\subsection{LIWC Analysis of Different Subsets}

We conduct use the LIWC word categories to analyze the linguistic features of different subsets in \cref{tab:liwc2,tab:liwc4}.

\begin{table*}[t]
    \centering
    \small
    \resizebox{\textwidth}{!}{\begin{tabular}{llccccc}
    \toprule
    Category & & \textbf{Score (All)} & \multicolumn{3}{c}{\textbf{Score (Subsets)}} \\
    \cline{4-6}
    & & & \textcolor{blue}{\textbf{Same Correct}} & \textcolor{red}{\textbf{Same Incorrect}} & \textcolor{violet}{\textbf{Diverse}} \\ \midrule
    Word Count \\ \midrule
    \multicolumn{3}{l}{\textbf{\textit{Summary Language Variables}}} \\
Words/Sentence &  &   138.33 $\pm$ 125.20   &   144.46 $\pm$ 126.86   &   126.85 $\pm$ 118.85   &   131.55 $\pm$ 125.40  \\
\multicolumn{3}{l}{	\textbf{\textit{Linguistic Dimensions}}} &  &  &  &  \\
{Total Function Words} & \{the, of, and, a, to\} & 59.92 $\pm$ 56.71  & 62.81 $\pm$ 57.73  & 54.66 $\pm$ 53.63  & 56.57 $\pm$ 56.09  \\
\hspace{1.5mm}{Total Pronouns} & \{that, this, we, which, it\} & 10.90 $\pm$ 12.20  & 11.61 $\pm$ 12.61  & 9.72 $\pm$ 11.45  & 9.96 $\pm$ 11.50  \\
\hspace{1.5mm}{Personal Pronouns} & \{we, our, they, them, us\} & 5.89 $\pm$ 7.49  & 6.34 $\pm$ 7.80  & 5.11 $\pm$ 7.01  & 5.32 $\pm$ 6.90  \\
\hspace{3mm}{1st Person Singular} & \{i, mine, my, im, me\} & 1.61 $\pm$ 2.78  & 1.75 $\pm$ 2.94  & 1.41 $\pm$ 2.60  & 1.42 $\pm$ 2.45  \\
\hspace{3mm}{1st Person Plural} & \{we, our, us, lets, ourselves\} & 1.37 $\pm$ 3.02  & 1.47 $\pm$ 3.13  & 1.17 $\pm$ 2.83  & 1.29 $\pm$ 2.86  \\
\hspace{3mm}{2nd Person} & \{you, your, u, ya, ye\} & 1.22 $\pm$ 2.11  & 1.27 $\pm$ 2.21  & 1.14 $\pm$ 1.96  & 1.12 $\pm$ 1.96  \\
\hspace{3mm}{3rd Person Singular} & \{his, her, he, she, him\} & 0.72 $\pm$ 2.12  & 0.83 $\pm$ 2.24  & 0.53 $\pm$ 2.02  & 0.61 $\pm$ 1.81  \\
\hspace{3mm}{3rd Person Plural} & \{they, them, themselves, their, theirs\} & 0.96 $\pm$ 1.63  & 1.02 $\pm$ 1.66  & 0.86 $\pm$ 1.54  & 0.88 $\pm$ 1.61  \\
\hspace{1.5mm}{Impersonal Pronouns} & \{that, this, which, it, these\} & 5.00 $\pm$ 5.54  & 5.26 $\pm$ 5.71  & 4.59 $\pm$ 5.21  & 4.63 $\pm$ 5.32  \\
\hspace{1.5mm}\blue{Articles} & \{the, a an\} & 9.51 $\pm$ 9.48  & 9.85 $\pm$ 9.58  & 8.84 $\pm$ 9.08  & 9.18 $\pm$ 9.57  \\
\hspace{1.5mm}{Prepositions} & \{of, to, in, for, with\} & 15.35 $\pm$ 15.50  & 16.05 $\pm$ 15.73  & 13.98 $\pm$ 14.83  & 14.62 $\pm$ 15.36  \\
\hspace{1.5mm}Auxiliary Verbs & \{is, are, be can, have\} & 11.34 $\pm$ 10.45  & 11.85 $\pm$ 10.60  & 10.47 $\pm$ 9.89  & 10.67 $\pm$ 10.45  \\
\hspace{1.5mm}{Common Adverbs} & \{such, also, when, only, where\} & 6.19 $\pm$ 6.22  & 6.48 $\pm$ 6.36  & 5.64 $\pm$ 5.81  & 5.84 $\pm$ 6.13  \\
\hspace{1.5mm}{Conjunctions} & \{and, as, or, also, but\} & 8.25 $\pm$ 7.92  & 8.60 $\pm$ 8.03  & 7.55 $\pm$ 7.41  & 7.91 $\pm$ 8.05  \\
\hspace{1.5mm}{Negations} & \{not, without, no, cannot, negative\} & 2.33 $\pm$ 2.69  & 2.48 $\pm$ 2.78  & 2.06 $\pm$ 2.46  & 2.17 $\pm$ 2.62  \\
\multicolumn{3}{l}{	\textbf{\textit{Other Grammar}}} &  &  &  &  \\
\hspace{1.5mm}{Common Verbs} & \{is, are, be, using, based\} & 19.85 $\pm$ 19.00  & 20.91 $\pm$ 19.39  & 17.97 $\pm$ 18.02  & 18.58 $\pm$ 18.53  \\
\hspace{1.5mm}{Common Adjectives} &  \{as, different, new, more, than\} & 5.49 $\pm$ 5.30  & 5.67 $\pm$ 5.30  & 5.13 $\pm$ 5.13  & 5.33 $\pm$ 5.46  \\
\hspace{1.5mm}{Comparisons} & \{as, different, more, than, most\} & 2.68 $\pm$ 3.07  & 2.79 $\pm$ 3.10  & 2.40 $\pm$ 2.87  & 2.64 $\pm$ 3.14  \\
\hspace{1.5mm}{Interrogatives} & \{which, when, where, how, whether\} & 1.28 $\pm$ 1.80  & 1.35 $\pm$ 1.87  & 1.16 $\pm$ 1.66  & 1.21 $\pm$ 1.72  \\
\hspace{1.5mm}Numbers & \{two, one, first, three, single\} & 0.91 $\pm$ 1.45  & 0.96 $\pm$ 1.48  & 0.78 $\pm$ 1.36  & 0.87 $\pm$ 1.43  \\
\hspace{1.5mm}Quantifiers & \{more, each, both, most, all\} & 2.60 $\pm$ 2.97  & 2.67 $\pm$ 2.97  & 2.46 $\pm$ 2.89  & 2.54 $\pm$ 3.03  \\
\multicolumn{3}{l}{	\textbf{\textit{Psychological Processes}}} &  &  &  &  \\
{Affective Processes} & \{well, important, problems, energy, problem\} & 4.98 $\pm$ 4.58  & 5.19 $\pm$ 4.62  & 4.62 $\pm$ 4.32  & 4.71 $\pm$ 4.68  \\
\hspace{1.5mm}{Positive Emotion} & \{well, important, energy, better, support\} & 3.59 $\pm$ 3.53  & 3.71 $\pm$ 3.59  & 3.40 $\pm$ 3.31  & 3.45 $\pm$ 3.54  \\
\hspace{1.5mm}Negative Emotion & \{problems, problem, low, critical, difficult\} & 1.26 $\pm$ 1.85  & 1.35 $\pm$ 1.91  & 1.09 $\pm$ 1.73  & 1.15 $\pm$ 1.74  \\
\hspace{3mm}Anxiety & \{uncertainty, pressure, uncertainties, risk, risks\} & 0.14 $\pm$ 0.44  & 0.15 $\pm$ 0.45  & 0.12 $\pm$ 0.38  & 0.13 $\pm$ 0.46  \\
\hspace{3mm}{Anger} & \{critical, attacks, argue, dominant, arguments\} & 0.27 $\pm$ 0.69  & 0.29 $\pm$ 0.72  & 0.21 $\pm$ 0.66  & 0.25 $\pm$ 0.63  \\
\hspace{3mm}Sadness & \{low, lower, failure, missing, suffer\} & 0.27 $\pm$ 0.60  & 0.28 $\pm$ 0.62  & 0.25 $\pm$ 0.57  & 0.25 $\pm$ 0.57  \\
{Social Processes} & \{we, our, provide, they, provides\} & 7.54 $\pm$ 9.52  & 8.14 $\pm$ 10.02  & 6.44 $\pm$ 8.62  & 6.86 $\pm$ 8.67  \\
\hspace{1.5mm}{Family} & \{family, families, parents, pregnancy, son\} & 0.15 $\pm$ 0.55  & 0.17 $\pm$ 0.60  & 0.12 $\pm$ 0.45  & 0.14 $\pm$ 0.49  \\
\hspace{1.5mm}{Friends} & \{contact, neighborhood, neighboring, neighbors, date\} & 0.33 $\pm$ 0.70  & 0.33 $\pm$ 0.70  & 0.31 $\pm$ 0.69  & 0.32 $\pm$ 0.70  \\
\hspace{1.5mm}{Female References} & \{female, her, women, females, she\} & 0.51 $\pm$ 1.75  & 0.58 $\pm$ 1.91  & 0.39 $\pm$ 1.51  & 0.42 $\pm$ 1.43  \\
\hspace{1.5mm}{Male References} & \{his, male, he, men, son\} & 0.48 $\pm$ 1.43  & 0.53 $\pm$ 1.51  & 0.38 $\pm$ 1.34  & 0.43 $\pm$ 1.25  \\
{{Cognitive Processes}} & \{using, based, or, used, results\} & 11.93 $\pm$ 11.88  & 12.50 $\pm$ 12.14  & 10.82 $\pm$ 10.89  & 11.35 $\pm$ 11.93  \\
\hspace{1.5mm}{{Insight}} & \{information, learning, analysis, knowledge, recognition\}  & 1.64 $\pm$ 2.20  & 1.72 $\pm$ 2.25  & 1.48 $\pm$ 2.09  & 1.53 $\pm$ 2.13  \\
\hspace{1.5mm}{Causation} & \{using, based, used, results, use\} & 1.30 $\pm$ 1.81  & 1.37 $\pm$ 1.83  & 1.16 $\pm$ 1.72  & 1.23 $\pm$ 1.80  \\
\hspace{1.5mm}{Discrepancy} & \{problems, problem, need, could, if\} & 1.80 $\pm$ 2.42  & 1.90 $\pm$ 2.52  & 1.60 $\pm$ 2.17  & 1.70 $\pm$ 2.32  \\
\hspace{1.5mm}Tentative & \{or, most, may, some, any\} & 2.77 $\pm$ 3.25  & 2.86 $\pm$ 3.30  & 2.58 $\pm$ 3.12  & 2.66 $\pm$ 3.22  \\
\hspace{1.5mm}{Certainty} & \{all, accuracy, specific, accurate, total\} & 1.51 $\pm$ 1.86  & 1.61 $\pm$ 1.91  & 1.32 $\pm$ 1.69  & 1.38 $\pm$ 1.85  \\
\hspace{1.5mm}{Differentiation} & \{or, different, not, than, other\} & 4.33 $\pm$ 4.58  & 4.51 $\pm$ 4.71  & 3.99 $\pm$ 4.18  & 4.16 $\pm$ 4.57  \\
Perceptual Processes & \{show, images, search, fuzzy, image\} & 2.68 $\pm$ 3.26  & 2.79 $\pm$ 3.30  & 2.44 $\pm$ 3.12  & 2.59 $\pm$ 3.26  \\
\hspace{1.5mm}See & \{show, images, search, image, shows\} & 0.87 $\pm$ 1.46  & 0.90 $\pm$ 1.46  & 0.80 $\pm$ 1.46  & 0.88 $\pm$ 1.47  \\
\hspace{1.5mm}{Hear} & \{noise, noisy, music, voice, speech\} & 0.53 $\pm$ 1.14  & 0.57 $\pm$ 1.21  & 0.46 $\pm$ 0.97  & 0.49 $\pm$ 1.08  \\
\hspace{1.5mm}\blue{Feel} & \{fuzzy, flexible, weight, weighted, hand\} & 0.57 $\pm$ 1.04  & 0.58 $\pm$ 1.05  & 0.53 $\pm$ 0.98  & 0.56 $\pm$ 1.07  \\
{Biological Processes} & \{clinical, expression, face, medical, physical\} & 3.64 $\pm$ 4.27  & 3.68 $\pm$ 4.24  & 3.52 $\pm$ 4.24  & 3.63 $\pm$ 4.40  \\
\hspace{1.5mm}Body & \{face, blood, hand, heart, neurons\} & 0.32 $\pm$ 0.81  & 0.34 $\pm$ 0.85  & 0.29 $\pm$ 0.74  & 0.31 $\pm$ 0.77  \\
\hspace{1.5mm}{Health} & \{clinical, medical, physical, health, diagnosis\} & 0.24 $\pm$ 0.73  & 0.27 $\pm$ 0.79  & 0.21 $\pm$ 0.64  & 0.21 $\pm$ 0.57  \\
\hspace{1.5mm}{Sexual} & \{prostate, pregnancy, sex, ovarian, arousal\} & 0.03 $\pm$ 0.21  & 0.03 $\pm$ 0.22  & 0.02 $\pm$ 0.18  & 0.03 $\pm$ 0.22  \\
\hspace{1.5mm}{Ingestion} & \{expression, water, weight, expressions, expressed\} & 2.94 $\pm$ 3.85  & 2.95 $\pm$ 3.83  & 2.90 $\pm$ 3.86  & 2.98 $\pm$ 3.92  \\
{Drives} & \{we, approach, our, first, over\} & 7.28 $\pm$ 7.69  & 7.68 $\pm$ 7.89  & 6.50 $\pm$ 7.03  & 6.87 $\pm$ 7.62  \\
\hspace{1.5mm}{Affiliation} & \{we, our, social, communication, interaction\} & 2.20 $\pm$ 3.67  & 2.33 $\pm$ 3.79  & 1.93 $\pm$ 3.38  & 2.08 $\pm$ 3.55  \\
\hspace{1.5mm}{Achievement} &  \{first, work, efficient, obtained, better\} & 1.26 $\pm$ 1.61  & 1.35 $\pm$ 1.65  & 1.13 $\pm$ 1.48  & 1.15 $\pm$ 1.59  \\
\hspace{1.5mm}{Power} & \{over, high, order, large, important\} & 1.91 $\pm$ 2.41  & 2.02 $\pm$ 2.46  & 1.66 $\pm$ 2.17  & 1.84 $\pm$ 2.47  \\
\hspace{1.5mm}{Reward} & \{approach, obtained, approaches, better, best\} & 2.13 $\pm$ 2.27  & 2.23 $\pm$ 2.33  & 1.97 $\pm$ 2.12  & 1.99 $\pm$ 2.25  \\
\hspace{1.5mm}Risk & \{problems, problem, security, difficult, lack\} & 0.40 $\pm$ 0.78  & 0.43 $\pm$ 0.82  & 0.35 $\pm$ 0.71  & 0.36 $\pm$ 0.71  \\
Time Orientations &  &  &  &  \\
\hspace{1.5mm}{Past Focus} & \{used, was, been, were, obtained\} & 8.20 $\pm$ 9.58  & 8.71 $\pm$ 9.79  & 7.33 $\pm$ 9.34  & 7.57 $\pm$ 9.06  \\
\hspace{1.5mm}Present Focus & \{is, are, be, can, have\} & 9.53 $\pm$ 9.41  & 9.96 $\pm$ 9.61  & 8.74 $\pm$ 8.60  & 9.06 $\pm$ 9.52  \\
\hspace{1.5mm}Future Focus & \{may, then, will, prediction, future\} & 1.01 $\pm$ 1.41  & 1.07 $\pm$ 1.45  & 0.94 $\pm$ 1.37  & 0.90 $\pm$ 1.35  \\
{Relativity} & \{in, on, at, approach, new\} & 14.93 $\pm$ 15.16  & 15.69 $\pm$ 15.51  & 13.58 $\pm$ 14.37  & 14.01 $\pm$ 14.73  \\
\hspace{1.5mm}Motion & \{approach, approaches, behavior, changes, increase\} & 2.44 $\pm$ 2.94  & 2.59 $\pm$ 3.03  & 2.15 $\pm$ 2.73  & 2.29 $\pm$ 2.86  \\
\hspace{1.5mm}{Space} & \{in, on, at, into, both\} & 8.19 $\pm$ 8.54  & 8.56 $\pm$ 8.71  & 7.59 $\pm$ 8.32  & 7.69 $\pm$ 8.19  \\
\hspace{1.5mm}Time & \{new, present, first, when, then\} & 4.63 $\pm$ 5.43  & 4.91 $\pm$ 5.63  & 4.14 $\pm$ 4.99  & 4.30 $\pm$ 5.20  \\
Personal Concerns &  &  &  &  \\
\hspace{1.5mm}{Work} & \{performance, learning, analysis, paper, applications\}& 1.40 $\pm$ 2.12  & 1.53 $\pm$ 2.27  & 1.20 $\pm$ 1.77  & 1.23 $\pm$ 1.97  \\
\hspace{1.5mm}Leisure & \{novel, expression, channels, videos, play\} & 1.43 $\pm$ 2.17  & 1.45 $\pm$ 2.22  & 1.45 $\pm$ 2.17  & 1.37 $\pm$ 2.03  \\
\hspace{1.5mm}Home & \{address, family, home, neighborhood, neighboring\} & 0.49 $\pm$ 1.21  & 0.51 $\pm$ 1.24  & 0.47 $\pm$ 1.23  & 0.45 $\pm$ 1.06  \\
\hspace{1.5mm}Money & \{investigate, cost, investigated, free, economic\} & 1.03 $\pm$ 1.69  & 1.08 $\pm$ 1.73  & 0.99 $\pm$ 1.71  & 0.94 $\pm$ 1.51  \\
\hspace{1.5mm}{Religion} & \{beliefs, moral, sacrificing, monkeys, agnostic\} & 0.04 $\pm$ 0.21  & 0.04 $\pm$ 0.22  & 0.03 $\pm$ 0.21  & 0.03 $\pm$ 0.19  \\
\hspace{1.5mm}{Death} & \{mortality, die, mortality, deaths, death\} & 0.04 $\pm$ 0.21  & 0.04 $\pm$ 0.20  & 0.03 $\pm$ 0.22  & 0.04 $\pm$ 0.22  \\
Informal Language & \{well, o, da, en, um\} & 0.48 $\pm$ 0.92  & 0.48 $\pm$ 0.93  & 0.49 $\pm$ 0.91  & 0.49 $\pm$ 0.89  \\
\hspace{1.5mm}Swear Words & \{retardation, dummy, screws, screw, retarded\} & 0.09 $\pm$ 0.36  & 0.09 $\pm$ 0.38  & 0.08 $\pm$ 0.35  & 0.08 $\pm$ 0.34  \\
\hspace{1.5mm}{Netspeak} & \{o, da, em, k, mm\} & 0.12 $\pm$ 0.44  & 0.12 $\pm$ 0.45  & 0.13 $\pm$ 0.44  & 0.12 $\pm$ 0.41  \\
\hspace{1.5mm}Assent & \{k, indeed, agree, absolutely, cool\} & 0.14 $\pm$ 0.42  & 0.14 $\pm$ 0.42  & 0.13 $\pm$ 0.41  & 0.15 $\pm$ 0.42  \\
\hspace{1.5mm}Nonfluencies & \{well, um, mm, er, ah\} & 0.10 $\pm$ 0.35  & 0.10 $\pm$ 0.34  & 0.11 $\pm$ 0.37  & 0.10 $\pm$ 0.36  \\
\hspace{1.5mm}Fillers & \{rrani, rranr\} & 0.02 $\pm$ 0.14  & 0.02 $\pm$ 0.15  & 0.01 $\pm$ 0.12  & 0.02 $\pm$ 0.14  \\
    \bottomrule
    \end{tabular}
    }
    \caption{LIWC analysis results on all Yelp data (\textbf{Score (All)}), and different subsets (\textcolor{blue}{\textbf{Same Correct}}, \textcolor{red}{\textbf{Same Incorrect}}, and \textcolor{violet}{\textbf{Diverse}}).}
    \label{tab:liwc2}
\end{table*}

\begin{table*}[t]
    \centering
    \small
    \resizebox{\textwidth}{!}{\begin{tabular}{llccccc}
    \toprule
    Category & & \textbf{Score (All \textcolor{violet}{\textbf{Diverse}})} & \multicolumn{3}{c}{\textbf{Score (Subsets)}} \\
    \cline{4-6}
    & & & \textbf{\textcolor{blue}{C1}$\ne$GT} & \textbf{\textcolor{red}{C2}$\ne$GT} &  \textbf{\textcolor{green}{C3}$\ne$GT} \\ \midrule
    Word Count \\ \midrule
    \multicolumn{3}{l}{\textbf{\textit{Summary Language Variables}}} \\
Words/Sentence &  & 131.55 $\pm$ 125.40  & 124.21 $\pm$ 121.95  & 128.16 $\pm$ 122.18  & 128.97 $\pm$ 127.07  \\
\multicolumn{3}{l}{	\textbf{\textit{Linguistic Dimensions}}} &  &  &  &  \\
{Total Function Words} & \{the, of, and, a, to\} & 56.57 $\pm$ 56.09  & 53.21 $\pm$ 54.10  & 55.27 $\pm$ 54.66  & 55.11 $\pm$ 56.56  \\
\hspace{1.5mm}{Total Pronouns} & \{that, this, we, which, it\} & 9.96 $\pm$ 11.50  & 9.39 $\pm$ 10.94  & 9.76 $\pm$ 11.35  & 9.71 $\pm$ 11.67  \\
\hspace{1.5mm}{Personal Pronouns} & \{we, our, they, them, us\} & 5.32 $\pm$ 6.90  & 5.01 $\pm$ 6.48  & 5.19 $\pm$ 6.80  & 5.19 $\pm$ 6.98  \\
\hspace{3mm}{1st Person Singular} & \{i, mine, my, im, me\} & 1.42 $\pm$ 2.45  & 1.33 $\pm$ 2.34  & 1.37 $\pm$ 2.41  & 1.40 $\pm$ 2.47  \\
\hspace{3mm}{1st Person Plural} & \{we, our, us, lets, ourselves\} & 1.29 $\pm$ 2.86  & 1.18 $\pm$ 2.74  & 1.29 $\pm$ 2.94  & 1.18 $\pm$ 2.72  \\
\hspace{3mm}{2nd Person} & \{you, your, u, ya, ye\} & 1.12 $\pm$ 1.96  & 1.12 $\pm$ 1.90  & 1.07 $\pm$ 1.89  & 1.14 $\pm$ 2.07  \\
\hspace{3mm}{3rd Person Singular} & \{his, her, he, she, him\} & 0.61 $\pm$ 1.81  & 0.58 $\pm$ 1.73  & 0.58 $\pm$ 1.75  & 0.63 $\pm$ 1.92  \\
\hspace{3mm}{3rd Person Plural} & \{they, them, themselves, their, theirs\} & 0.88 $\pm$ 1.61  & 0.79 $\pm$ 1.35  & 0.89 $\pm$ 1.68  & 0.84 $\pm$ 1.60  \\
\hspace{1.5mm}{Impersonal Pronouns} & \{that, this, which, it, these\} & 4.63 $\pm$ 5.32  & 4.36 $\pm$ 5.13  & 4.55 $\pm$ 5.33  & 4.51 $\pm$ 5.35  \\
\hspace{1.5mm}\blue{Articles} & \{the, a an\} & 9.18 $\pm$ 9.57  & 8.60 $\pm$ 9.28  & 8.99 $\pm$ 9.34  & 9.05 $\pm$ 9.74  \\
\hspace{1.5mm}{Prepositions} & \{of, to, in, for, with\}& 14.62 $\pm$ 15.36  & 13.68 $\pm$ 14.63  & 14.37 $\pm$ 15.22  & 14.37 $\pm$ 15.54  \\
\hspace{1.5mm}Auxiliary Verbs & \{is, are, be can, have\}& 10.67 $\pm$ 10.45  & 10.15 $\pm$ 10.19  & 10.31 $\pm$ 10.08  & 10.28 $\pm$ 10.47  \\
\hspace{1.5mm}{Common Adverbs} & \{such, also, when, only, where\} & 5.84 $\pm$ 6.13  & 5.48 $\pm$ 5.85  & 5.76 $\pm$ 6.14  & 5.57 $\pm$ 5.82  \\
\hspace{1.5mm}{Conjunctions} & \{and, as, or, also, but\} & 7.91 $\pm$ 8.05  & 7.41 $\pm$ 7.89  & 7.69 $\pm$ 7.62  & 7.68 $\pm$ 8.04  \\
\hspace{1.5mm}{Negations} & \{not, without, no, cannot, negative\} & 2.17 $\pm$ 2.62  & 2.08 $\pm$ 2.59  & 2.11 $\pm$ 2.51  & 2.12 $\pm$ 2.66  \\
\multicolumn{3}{l}{	\textbf{\textit{Other Grammar}}} &  &  &  &  \\
\hspace{1.5mm}{Common Verbs} & \{is, are, be, using, based\} & 18.58 $\pm$ 18.53  & 17.55 $\pm$ 17.90  & 18.08 $\pm$ 17.85  & 18.07 $\pm$ 18.93  \\
\hspace{1.5mm}{Common Adjectives} &  \{as, different, new, more, than\} & 5.33 $\pm$ 5.46  & 4.99 $\pm$ 5.16  & 5.25 $\pm$ 5.50  & 5.20 $\pm$ 5.41  \\
\hspace{1.5mm}{Comparisons} & \{as, different, more, than, most\} & 2.64 $\pm$ 3.14  & 2.47 $\pm$ 3.01  & 2.64 $\pm$ 3.13  & 2.52 $\pm$ 3.08  \\
\hspace{1.5mm}{Interrogatives} & \{which, when, where, how, whether\} & 1.21 $\pm$ 1.72  & 1.12 $\pm$ 1.62  & 1.18 $\pm$ 1.76  & 1.20 $\pm$ 1.72  \\
\hspace{1.5mm}Numbers & \{two, one, first, three, single\} & 0.87 $\pm$ 1.43  & 0.85 $\pm$ 1.49  & 0.87 $\pm$ 1.42  & 0.86 $\pm$ 1.41  \\
\hspace{1.5mm}Quantifiers & \{more, each, both, most, all\} & 2.54 $\pm$ 3.03  & 2.44 $\pm$ 2.92  & 2.48 $\pm$ 3.05  & 2.42 $\pm$ 2.89  \\
\multicolumn{3}{l}{	\textbf{\textit{Psychological Processes}}} &  &  &  &  \\
{Affective Processes} & \{well, important, problems, energy, problem\} & 4.71 $\pm$ 4.68  & 4.35 $\pm$ 4.30  & 4.62 $\pm$ 4.64  & 4.54 $\pm$ 4.57  \\
\hspace{1.5mm}{Positive Emotion} & \{well, important, energy, better, support\} & 3.45 $\pm$ 3.54  & 3.14 $\pm$ 3.30  & 3.38 $\pm$ 3.52  & 3.33 $\pm$ 3.47  \\
\hspace{1.5mm}Negative Emotion & \{problems, problem, low, critical, difficult\}& 1.15 $\pm$ 1.74  & 1.10 $\pm$ 1.63  & 1.11 $\pm$ 1.65  & 1.10 $\pm$ 1.69  \\
\hspace{3mm}Anxiety & \{uncertainty, pressure, uncertainties, risk, risks\} & 0.13 $\pm$ 0.46  & 0.12 $\pm$ 0.40  & 0.12 $\pm$ 0.38  & 0.14 $\pm$ 0.50  \\
\hspace{3mm}{Anger} & \{critical, attacks, argue, dominant, arguments\} & 0.25 $\pm$ 0.63  & 0.23 $\pm$ 0.58  & 0.23 $\pm$ 0.59  & 0.22 $\pm$ 0.60  \\
\hspace{3mm}Sadness & \{low, lower, failure, missing, suffer\} & 0.25 $\pm$ 0.57  & 0.26 $\pm$ 0.60  & 0.23 $\pm$ 0.52  & 0.24 $\pm$ 0.57  \\
{Social Processes} & \{we, our, provide, they, provides\} & 6.86 $\pm$ 8.67  & 6.55 $\pm$ 8.45  & 6.66 $\pm$ 8.34  & 6.79 $\pm$ 8.99  \\
\hspace{1.5mm}{Family} & \{family, families, parents, pregnancy, son\} & 0.14 $\pm$ 0.49  & 0.13 $\pm$ 0.44  & 0.13 $\pm$ 0.44  & 0.15 $\pm$ 0.54  \\
\hspace{1.5mm}{Friends} & \{contact, neighborhood, neighboring, neighbors, date\} & 0.32 $\pm$ 0.70  & 0.34 $\pm$ 0.73  & 0.30 $\pm$ 0.68  & 0.31 $\pm$ 0.69  \\
\hspace{1.5mm}{Female References} & \{female, her, women, females, she\} & 0.42 $\pm$ 1.43  & 0.40 $\pm$ 1.35  & 0.38 $\pm$ 1.39  & 0.43 $\pm$ 1.52  \\
\hspace{1.5mm}{Male References} & \{his, male, he, men, son\} & 0.43 $\pm$ 1.25  & 0.43 $\pm$ 1.20  & 0.41 $\pm$ 1.26  & 0.46 $\pm$ 1.34  \\
{{Cognitive Processes}} & \{using, based, or, used, results\}& 11.35 $\pm$ 11.93  & 10.71 $\pm$ 11.39  & 11.09 $\pm$ 11.85  & 10.98 $\pm$ 11.73  \\
\hspace{1.5mm}{{Insight}} & \{information, learning, analysis, knowledge, recognition\}  & 1.53 $\pm$ 2.13  & 1.40 $\pm$ 1.97  & 1.53 $\pm$ 2.09  & 1.50 $\pm$ 2.24  \\
\hspace{1.5mm}{Causation} & \{using, based, used, results, use\} & 1.23 $\pm$ 1.80  & 1.13 $\pm$ 1.69  & 1.21 $\pm$ 1.76  & 1.20 $\pm$ 1.76  \\
\hspace{1.5mm}{Discrepancy} & \{problems, problem, need, could, if\} & 1.70 $\pm$ 2.32  & 1.59 $\pm$ 2.22  & 1.66 $\pm$ 2.26  & 1.65 $\pm$ 2.25  \\
\hspace{1.5mm}Tentative & \{or, most, may, some, any\} & 2.66 $\pm$ 3.22  & 2.51 $\pm$ 3.09  & 2.60 $\pm$ 3.19  & 2.54 $\pm$ 3.11  \\
\hspace{1.5mm}{Certainty} & \{all, accuracy, specific, accurate, total\} & 1.38 $\pm$ 1.85  & 1.37 $\pm$ 1.82  & 1.32 $\pm$ 1.83  & 1.34 $\pm$ 1.80  \\
\hspace{1.5mm}{Differentiation} & \{or, different, not, than, other\} & 4.16 $\pm$ 4.57  & 3.93 $\pm$ 4.52  & 4.08 $\pm$ 4.48  & 4.02 $\pm$ 4.47  \\
Perceptual Processes & \{show, images, search, fuzzy, image\} & 2.59 $\pm$ 3.26  & 2.41 $\pm$ 3.00  & 2.52 $\pm$ 3.27  & 2.52 $\pm$ 3.25  \\
\hspace{1.5mm}See & \{show, images, search, image, shows\} & 0.88 $\pm$ 1.47  & 0.82 $\pm$ 1.33  & 0.86 $\pm$ 1.54  & 0.85 $\pm$ 1.50  \\
\hspace{1.5mm}{Hear} & \{noise, noisy, music, voice, speech\} & 0.49 $\pm$ 1.08  & 0.46 $\pm$ 0.97  & 0.47 $\pm$ 1.14  & 0.49 $\pm$ 1.03  \\
\hspace{1.5mm}\blue{Feel} & \{fuzzy, flexible, weight, weighted, hand\} & 0.56 $\pm$ 1.07  & 0.50 $\pm$ 0.91  & 0.55 $\pm$ 1.09  & 0.55 $\pm$ 1.12  \\
{Biological Processes} & \{clinical, expression, face, medical, physical\} & 3.63 $\pm$ 4.40  & 3.36 $\pm$ 4.20  & 3.45 $\pm$ 4.22  & 3.49 $\pm$ 4.35  \\
\hspace{1.5mm}Body & \{face, blood, hand, heart, neurons\} & 0.31 $\pm$ 0.77  & 0.28 $\pm$ 0.75  & 0.28 $\pm$ 0.71  & 0.30 $\pm$ 0.76  \\
\hspace{1.5mm}{Health} & \{clinical, medical, physical, health, diagnosis\} & 0.21 $\pm$ 0.57  & 0.19 $\pm$ 0.52  & 0.21 $\pm$ 0.55  & 0.22 $\pm$ 0.58  \\
\hspace{1.5mm}{Sexual} & \{prostate, pregnancy, sex, ovarian, arousal\} & 0.03 $\pm$ 0.22  & 0.04 $\pm$ 0.27  & 0.03 $\pm$ 0.21  & 0.03 $\pm$ 0.25  \\
\hspace{1.5mm}{Ingestion} & \{expression, water, weight, expressions, expressed\} & 2.98 $\pm$ 3.92  & 2.76 $\pm$ 3.78  & 2.85 $\pm$ 3.79  & 2.83 $\pm$ 3.89  \\
{Drives} & \{we, approach, our, first, over\} & 6.87 $\pm$ 7.62  & 6.38 $\pm$ 7.28  & 6.79 $\pm$ 7.59  & 6.75 $\pm$ 7.65  \\
\hspace{1.5mm}{Affiliation} & \{we, our, social, communication, interaction\} & 2.08 $\pm$ 3.55  & 1.94 $\pm$ 3.44  & 2.06 $\pm$ 3.57  & 1.97 $\pm$ 3.38  \\
\hspace{1.5mm}{Achievement} &  \{first, work, efficient, obtained, better\} & 1.15 $\pm$ 1.59  & 1.07 $\pm$ 1.50  & 1.13 $\pm$ 1.56  & 1.14 $\pm$ 1.63  \\
\hspace{1.5mm}{Power} & \{over, high, order, large, important\} & 1.84 $\pm$ 2.47  & 1.65 $\pm$ 2.18  & 1.79 $\pm$ 2.49  & 1.90 $\pm$ 2.62  \\
\hspace{1.5mm}{Reward} & \{approach, obtained, approaches, better, best\} & 1.99 $\pm$ 2.25  & 1.87 $\pm$ 2.16  & 1.96 $\pm$ 2.24  & 1.92 $\pm$ 2.24  \\
\hspace{1.5mm}Risk & \{problems, problem, security, difficult, lack\} & 0.36 $\pm$ 0.71  & 0.36 $\pm$ 0.71  & 0.39 $\pm$ 0.74  & 0.37 $\pm$ 0.73  \\
Time Orientations &  &  &  &  \\
\hspace{1.5mm}{Past Focus} & \{used, was, been, were, obtained\} & 7.57 $\pm$ 9.06  & 7.16 $\pm$ 8.95  & 7.28 $\pm$ 8.78  & 7.28 $\pm$ 9.12  \\
\hspace{1.5mm}Present Focus & \{is, are, be, can, have\} & 9.06 $\pm$ 9.52  & 8.56 $\pm$ 9.04  & 8.94 $\pm$ 8.98  & 8.87 $\pm$ 9.96  \\
\hspace{1.5mm}Future Focus & \{may, then, will, prediction, future\} & 0.90 $\pm$ 1.35  & 0.90 $\pm$ 1.34  & 0.89 $\pm$ 1.29  & 0.91 $\pm$ 1.41  \\
{Relativity} & \{in, on, at, approach, new\} & 14.01 $\pm$ 14.73  & 13.21 $\pm$ 14.12  & 13.79 $\pm$ 14.58  & 13.78 $\pm$ 15.10  \\
\hspace{1.5mm}Motion & \{approach, approaches, behavior, changes, increase\} & 2.29 $\pm$ 2.86  & 2.22 $\pm$ 2.87  & 2.23 $\pm$ 2.77  & 2.29 $\pm$ 3.07  \\
\hspace{1.5mm}{Space} & \{in, on, at, into, both\} & 7.69 $\pm$ 8.19  & 7.20 $\pm$ 7.86  & 7.60 $\pm$ 8.24  & 7.52 $\pm$ 8.19  \\
\hspace{1.5mm}Time & \{new, present, first, when, then\} & 4.30 $\pm$ 5.20  & 4.05 $\pm$ 4.91  & 4.22 $\pm$ 4.97  & 4.23 $\pm$ 5.35  \\
Personal Concerns &  &  &  &  \\
\hspace{1.5mm}{Work} & \{performance, learning, analysis, paper, applications\}& 1.23 $\pm$ 1.97  & 1.14 $\pm$ 1.85  & 1.19 $\pm$ 1.85  & 1.27 $\pm$ 2.15  \\
\hspace{1.5mm}Leisure & \{novel, expression, channels, videos, play\} & 1.37 $\pm$ 2.03  & 1.34 $\pm$ 1.99  & 1.32 $\pm$ 1.92  & 1.31 $\pm$ 1.97  \\
\hspace{1.5mm}Home & \{address, family, home, neighborhood, neighboring\} & 0.45 $\pm$ 1.06  & 0.42 $\pm$ 1.03  & 0.47 $\pm$ 1.09  & 0.43 $\pm$ 1.04  \\
\hspace{1.5mm}Money & \{investigate, cost, investigated, free, economic\} & 0.94 $\pm$ 1.51  & 0.88 $\pm$ 1.48  & 0.94 $\pm$ 1.49  & 0.89 $\pm$ 1.50  \\
\hspace{1.5mm}{Religion} & \{beliefs, moral, sacrificing, monkeys, agnostic\} & 0.03 $\pm$ 0.19  & 0.03 $\pm$ 0.21  & 0.03 $\pm$ 0.19  & 0.03 $\pm$ 0.19  \\
\hspace{1.5mm}{Death} & \{mortality, die, mortality, deaths, death\} & 0.04 $\pm$ 0.22  & 0.04 $\pm$ 0.22  & 0.04 $\pm$ 0.22  & 0.04 $\pm$ 0.24  \\
Informal Language & \{well, o, da, en, um\} & 0.49 $\pm$ 0.89  & 0.45 $\pm$ 0.86  & 0.48 $\pm$ 0.90  & 0.48 $\pm$ 0.86  \\
\hspace{1.5mm}Swear Words & \{retardation, dummy, screws, screw, retarded\} & 0.08 $\pm$ 0.34  & 0.07 $\pm$ 0.30  & 0.08 $\pm$ 0.35  & 0.08 $\pm$ 0.33  \\
\hspace{1.5mm}{Netspeak} & \{o, da, em, k, mm\} & 0.12 $\pm$ 0.41  & 0.11 $\pm$ 0.40  & 0.11 $\pm$ 0.41  & 0.13 $\pm$ 0.43  \\
\hspace{1.5mm}Assent & \{k, indeed, agree, absolutely, cool\} & 0.15 $\pm$ 0.42  & 0.13 $\pm$ 0.42  & 0.15 $\pm$ 0.42  & 0.14 $\pm$ 0.39  \\
\hspace{1.5mm}Nonfluencies & \{well, um, mm, er, ah\} & 0.10 $\pm$ 0.36  & 0.09 $\pm$ 0.32  & 0.11 $\pm$ 0.37  & 0.10 $\pm$ 0.34  \\
\hspace{1.5mm}Fillers & \{rrani, rranr\} & 0.02 $\pm$ 0.14  & 0.02 $\pm$ 0.14  & 0.02 $\pm$ 0.14  & 0.02 $\pm$ 0.14  \\
    \bottomrule
    \end{tabular}
    }
    \caption{LIWC analysis results on all diverse samples (\textbf{Score (All \textcolor{violet}{\textbf{Diverse}})}) copied from \cref{tab:liwc2}, and three types of subsets in the diverse set (\textbf{\textcolor{blue}{C1}$\ne$GT}, \textbf{\textcolor{red}{C2}$\ne$GT}, and \textbf{\textcolor{green}{C3}$\ne$GT}).
    }
    \label{tab:liwc4}
\end{table*}

\end{document}